\documentclass[11pt]{article}

\usepackage[final]{acl}

\usepackage{amsmath}
\usepackage{wasysym}
\usepackage{times}
\usepackage{latexsym}
\usepackage[many]{tcolorbox}
\usepackage[T1]{fontenc}
\usepackage{inconsolata}
\usepackage{graphicx}
\usepackage{microtype}
\usepackage{amssymb}
\usepackage{mathtools}
\usepackage{amsthm}
\usepackage{multirow}
\usepackage{makecell}
\usepackage{booktabs}
\usepackage{array}
\usepackage{longtable}
\usepackage{subcaption}
\usepackage{fontawesome}
\usepackage{centernot}

\theoremstyle{plain}

\theoremstyle{definition}

\theoremstyle{remark}

\newtcolorbox{prompt}[2][]{
    colback=white,
    colframe=gray!45,
    fonttitle=\bfseries,
    coltitle=black,
    sharp corners,
    title=#2,
    #1
}

\tcbset{
    promptstyle/.style={
        enhanced,
        colback=white,
        colframe=black,
        colbacktitle=gray!20,
        width=0.98\linewidth, 
        coltitle=black,
        rounded corners,
        sharp corners=north,
        boxrule=0.5pt,
        drop shadow=black!50!white,
        attach boxed title to top left={
            xshift=-2mm,
            yshift=-2mm
        },
        title code={%
            \begin{minipage}{0.8\linewidth}
                \centering\thetitle
            \end{minipage}
        },
        boxed title style={
            rounded corners,
            size=small,
            colback=gray!20,
        },
        fonttitle=\normalsize\bfseries,
        before upper={\parindent15pt}
    }
}

\newtcolorbox{promptbox}[1][]{
    promptstyle,
    title=Prompt,
    #1
}

\newcommand{\dataname}[0]{\textsc{KodCode~}}
\newcommand{\datanamen}[0]{\textsc{KodCode}}
\usepackage{xcolor}

\newcommand{\github}{\raisebox{-1.5pt}{\includegraphics[height=1em]{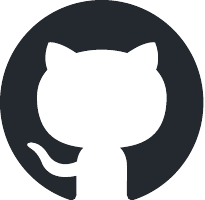}}}
\newcommand{\huggingface}{\raisebox{-1.5pt}{\includegraphics[height=1em]{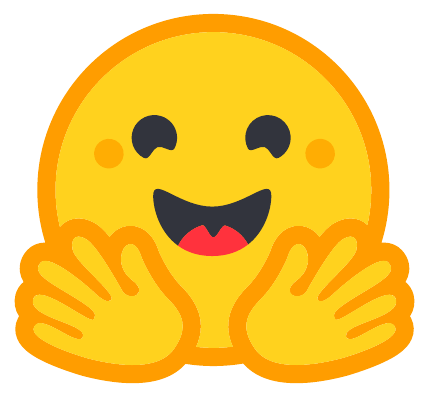}}}

\usepackage{listings}
\definecolor{codegreen}{rgb}{0,0.6,0}
\definecolor{codegray}{rgb}{0.5,0.5,0.5}
\definecolor{codepurple}{rgb}{0.58,0,0.82}
\definecolor{backcolour}{rgb}{0.95,0.95,0.92}

\lstdefinestyle{mystyle}{
    backgroundcolor=\color{white},
    commentstyle=\color{codegreen},
    keywordstyle=\color{magenta},
    numberstyle=\tiny\color{codegray},
    stringstyle=\color{codepurple},
    basicstyle=\ttfamily\footnotesize,
    breakatwhitespace=false,         
    breaklines=true,                 
    captionpos=b,                    
    keepspaces=true,                 
    numbers=none,
    numbersep=5pt,                  
    showspaces=false,                
    showstringspaces=false,
    showtabs=false,                  
    tabsize=2
}
\title{\dataname: A Diverse, Challenging, and Verifiable \\ Synthetic Dataset for Coding}

\author{%
  \textbf{Zhangchen Xu}\textsuperscript{$\clubsuit$\footnotemark[1]}\;\;%
  \textbf{Yang Liu}\textsuperscript{$\diamondsuit$}\;\;%
  \textbf{Yueqin Yin}\textsuperscript{$\spadesuit$\footnotemark[1]}\\[1ex]%
  \textbf{Mingyuan Zhou}\textsuperscript{$\spadesuit$}\;\;%
  \textbf{Radha Poovendran}\textsuperscript{$\clubsuit$}\\[1ex]%
  \textsuperscript{$\diamondsuit$}Microsoft GenAI \; 
  \textsuperscript{$\clubsuit$}University of Washington \; 
  \textsuperscript{$\spadesuit$}The University of Texas at Austin\\[1ex]%
  \normalsize{\texttt{\{zxu9,rp3\}@uw.edu, yaliu10@microsoft.com, \{yueqin.yin, mingyuan.zhou\}@utexas.edu}}\\[1ex]%
  {\github\ \texttt{\url{https://kodcode-ai.github.io}}}\quad
  {\huggingface\ \texttt{\url{https://huggingface.co/KodCode}}}
}

\begin{document}
\maketitle
\footnotetext[1]{Work done during internship at Microsoft GenAI.}

\begin{abstract}

We introduce \datanamen, a synthetic dataset that addresses the persistent challenge of acquiring high-quality, verifiable training data across diverse difficulties and domains for training Large Language Models for coding.
Existing code-focused resources typically fail to ensure either the breadth of coverage (e.g., spanning simple coding tasks to advanced algorithmic problems) or verifiable correctness (e.g., unit tests).
In contrast, \datanamen~comprises question–solution–test triplets that are systematically validated via a self-verification procedure.
Our pipeline begins by synthesizing a broad range of coding questions, then generates solutions and test cases with additional attempts allocated to challenging problems.
Finally, post-training data synthesis is done by rewriting questions into diverse formats and generating responses under a test-based reject sampling procedure from a reasoning model (DeepSeek R1). 
This pipeline yields a large-scale, robust and diverse coding dataset. \datanamen~is suitable for supervised fine-tuning and the paired unit tests also provide great potential for RL tuning. Fine-tuning experiments on coding benchmarks (HumanEval(+), MBPP(+), BigCodeBench, and LiveCodeBench) demonstrate that \datanamen-tuned models achieve state-of-the-art performance, surpassing models like Qwen2.5-Coder-32B-Instruct and DeepSeek-R1-Distill-Llama-70B.

\end{abstract}

\section{Introduction}

Recent advances in Large Language Models (LLMs) for coding such as Qwen2.5-Coder \cite{hui2024qwen25coder}, Deepseek Coder \cite{guo2024deepseek}, and OpenCoder \cite{Huang2024OpenCoderTO} have demonstrated remarkable capabilities in programming tasks. These models excel at function writing \cite{chen2021evaluating}, debugging \cite{zhong2024debuglikehumanlarge}, issue resolution \cite{zhang2024systematicliteraturereviewlarge}, and agent system enhancement \cite{zhang2024codeagentenhancingcodegeneration}, fundamentally transforming software development practices \cite{qian2024chatdevcommunicativeagentssoftware, hou2024largelanguagemodelssoftware}.

Ideally, training high-performing coding LLMs requires high-quality data with verified solutions and test cases for post-training stages, including supervised fine-tuning (SFT) and reinforcement learning (RL) \cite{deepseekr1, sky_t1_2025, hui2024qwen25coder, wei2024selfcodealignselfalignmentcodegeneration}. While human-curated coding datasets like TACO \cite{li2023taco}, APPS \cite{hendrycksapps2021}, and CodeContests \cite{li2022competition} offer high quality questions, canonical solutions, and tests, their limited scale constrains model training. Synthetic datasets have emerged as an alternative \cite{wang-etal-2023-self-instruct, long2024llms}, but often lack diversity \cite{xu2024magpie}, sufficient complexity \cite{luo2023wizardcoder}, and reliable response verification \cite{lei2024autocoder}.

\begin{table}[t]
    \centering
    \resizebox{0.48\textwidth}{!}{
    \begin{tabular}{@{}l ccccc@{}}
    \toprule
    Dataset Name & \#Problems & Diversity & Difficulty & \makecell{Unit\\Test} & \makecell{Verified\\Solution} \\ 
    \midrule
    APPS \citep{hendrycksapps2021} & 10K & High & High & \scalebox{1.5}{$\bullet$} & \scalebox{1.5}{$\bullet$} \\
    CodeContests \citep{li2022competition} & 13K & High & High & \scalebox{1.5}{$\bullet$} & \scalebox{1.5}{$\bullet$} \\
    TACO \citep{li2023taco} & 26K & High & High & \scalebox{1.5}{$\bullet$} & \scalebox{1.5}{$\bullet$} \\
    \midrule
    Code Alpaca \citep{codealpaca} & 20K & Low & Low & \scalebox{1.5}{$\circ$} & \scalebox{1.5}{$\circ$} \\
    SelfCodeAlign \citep{wei2024selfcodealignselfalignmentcodegeneration} & 50K & Mid & Low & \scalebox{1.5}{$\circ$} & \scalebox{1.5}{$\bullet$} \\
    OSS Instruct \citep{wei2024magicoder} & 75K & Mid & Mid & \scalebox{1.5}{$\circ$} & \scalebox{1.5}{$\circ$} \\
    AceCoder \citep{AceCoder} & 87K & Mid & Mid & \scalebox{1.5}{$\bullet$} & \scalebox{1.5}{$\circ$} \\
    Evol Instruct \citep{luo2023wizardcoder} & 111K & Low & Mid & \scalebox{1.5}{$\circ$} & \scalebox{1.5}{$\circ$} \\
    Educational Instruct \citep{Huang2024OpenCoderTO} & 118K & Low & Low & \scalebox{1.5}{$\bullet$} & \scalebox{1.5}{$\bullet$} \\
    Package Instruct \citep{Huang2024OpenCoderTO} & 171K & Mid & Mid & \scalebox{1.5}{$\circ$} & \scalebox{1.5}{$\circ$} \\
    \midrule
    \bf \dataname{}-V1 & \textbf{447K} & High & Mix & \scalebox{1.5}{$\bullet$} & \scalebox{1.5}{$\bullet$} \\
    \bottomrule
    \end{tabular}
    }
    \caption{Comparison of \datanamen~with existing code datasets for LLM post-training. The first three rows show human-curated datasets, while the remaining rows represent synthetic datasets. \datanamen~offers three difficulty labels (e.g., ``easy'', ``medium'', and ``hard''), which we denote as ``Mix''.}
    \vspace{-1em}
    \label{tab:compare with baselines}
\end{table}

\begin{figure*}[t!]
    \centering
    \includegraphics[width=1.0\linewidth]{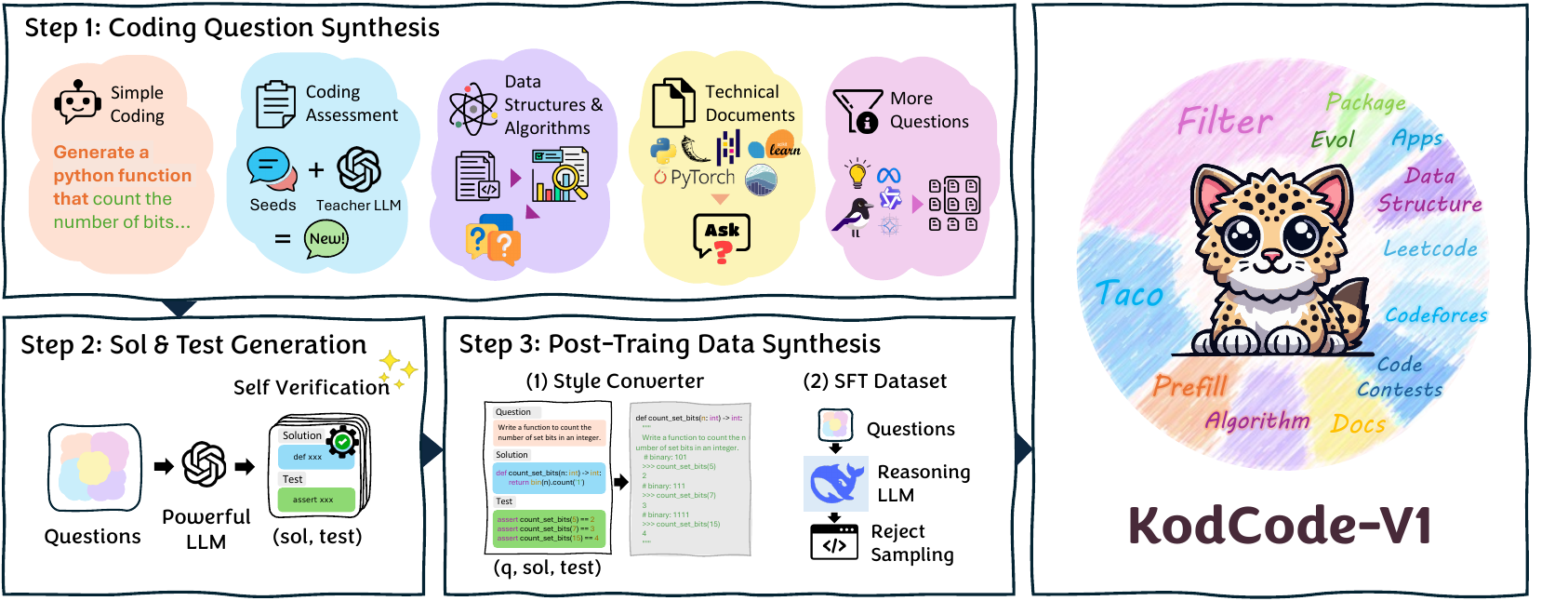}
    \caption{This figure demonstrates the pipeline for generating \datanamen-V1. Our approach follows a three-step pipeline: \textit{Coding Question Synthesis}, \textit{Solution \& Test Generation}, and \textit{Post-training Data Synthesis}. The final \datanamen-V1 dataset contains 447K verified question-solution-test triplets. The distribution of each subset is demonstrated on the right.}
    \vspace{-1em}
    \label{fig: pipeline}
\end{figure*}

In this paper, we bridge this gap by introducing \datanamen-V1, hereafter referred to as \datanamen, a synthetic dataset consisting of 447K coding questions with verified solutions and unit tests. Our approach starts from synthesizing coding questions from 12 sources using five distinct methods to ensure diversity and complexity. In solution \& test generation step, we generate unit tests along with the solution and execute the unit test to verify the correctness of the solution.
This \emph{self-verification} mechanism not only  ensures each solution is functionally correct but also offers verifiable correctness by providing explicitly curated unit tests.
In addition, for challenging questions whose solutions fail the self-verification, we allocate additional attempts rather than discarding them, ensuring challenging questions are not filtered out during reject sampling. We further enhance the dataset for post-training by rewriting questions into diverse formats and generating chain-of-thought (CoT) responses using DeepSeek-R1 \cite{deepseekr1} under a test-based reject sampling procedure.

We conduct comprehensive analyses of our coding data generation pipeline. First, we evaluate our self-verification mechanism by testing solutions against human-written unit tests from the MBPP validation dataset. 
Our experiments show that error rate remains below 2.5\%, demonstrating the effectiveness of our self-verification mechanism.
We then examine the benefits of scaled computation for challenging coding questions. Finally, we perform statistical analyses of \datanamen's token length, diversity, difficulty distribution, and potential contamination with existing benchmarks.

Furthermore, to validate \datanamen's effectiveness for code LLM post-training, we evaluate models using supervised SFT and RL across standard benchmarks: HumanEval(+), MBPP(+), BigCodeBench, and LiveCodeBench. Our experimental results show that \datanamen-fine-tuned models achieve state-of-the-art performance, surpassing other open-source models such as Qwen2.5-Coder-32B-Instruct and DeepSeek-R1-Distill-Llama-70B in most of the benchmarks. 



We hope our open-source dataset and models will help the community develop more capable coding assistants. We believe \datanamen~will advance current SFT and RL post-training pipelines for code generation models, pushing the boundaries of LLMs in coding tasks.

\section{\dataname: Synthesizing Diverse, Challenging, and Verifiable Correct Post-Training Data for Code}



As illustrated in Figure \ref{fig: pipeline}, our approach follows a three-step pipeline: \textit{Coding Question Synthesis}, \textit{Solution \& Test Generation}, and \textit{Post-training Data Synthesis}. Generally, we begin by synthesizing diverse coding questions $q$ through a combination of prompt engineering and LLM-based augmentation. Next, we leverage a self-verification process to create high-quality solutions and test cases $(sol, test)$ while offering verifiable correctness. Finally, to provide high-quality post-training data, we diversify the generated synthetic questions by rewriting them into different formats, and generate responses by prompting a reasoning model (i.e., DeepSeek-R1) with a test-based reject sampling procedure.

\subsection{Step 1: Coding Question Synthesis} \label{sec:step 1}

To generate challenging coding questions with broad coverage, we developed 12 distinct subsets spanning various domains (from algorithmic to package-specific knowledge) and difficulty levels (from basic coding exercises to interview and competitive programming challenges). Below, we elaborate on the pipeline used to construct each subset.


\textbf{Simple Coding Questions.} To generate simple coding questions, we extend the \textsc{Magpie} framework \cite{xu2024magpie} and introduce \textbf{\textsc{Magpie}-Prefill}. This approach generates simple coding questions by prefilling the user message in the chat template with a pre-defined suffix (e.g., \textit{``Write a Python function that''}), and leverages \textit{Qwen2.5-Coder-7B-Instruct} to complete the remaining part of the user query. This method efficiently generates diverse questions focused on function implementation and basic Python programming tasks. We name this subset \textbf{Prefill}. The complete prompt template is provided in Appendix \ref{appendix: magpie-prefill template}.



\textbf{Coding Assessment Questions.} To synthesize diverse coding assessment questions, we leverage existing human-written coding assessment datasets as seed corpora. To expand these datasets, we employ \textit{GPT-4o-0513} as a teacher LLM, prompting it to act as an expert programming instructor. The model analyzes the structure, complexity, and knowledge requirements of seed questions and subsequently generates new assessment questions that maintain consistency in difficulty and scope. 

The seed datasets utilized include LeetCode \cite{hartford2023leetcode}, Codeforces \cite{jur1cek2023codeforces}, APPS training subset \cite{hendrycksapps2021}, TACO training subset \cite{li2023taco}, and Code Contests \cite{li2022competition}. The respective question subsets are named \textbf{LeetCode}, \textbf{Codeforces}, \textbf{APPS}, \textbf{Taco}, and \textbf{Code Contests}. The complete prompt templates are shown in Appendix \ref{appendix: code assessment template}.

\textbf{Data Structures and Algorithms.} While coding assessments typically focus on specific programming concepts, they do not fully encompass Data Structures and Algorithms (DSA) knowledge. To bridge this gap, we convert Python DSA knowledge into assessment questions by uniformly sampling a collection of DSA code snippets \cite{thealgorithms2023python, keon2017pythonic}. We prompt LLM to first perform a systematic analysis of DSA snippets (addressing core components, complexity, and implementation challenges), then craft questions that test foundational understanding to avoid direct code replication. We denote the two subsets generated using this method as \textbf{Algorithm} and \textbf{Data Structure}. The corresponding prompt template can be found in Appendix \ref{appendix: algorithm template}. 

\color{black}
\textbf{Technical Documentations.} 
Given that users frequently ask package-related questions, we developed an additional subset called \textbf{Docs}, which transforms technical documentation from popular Python libraries—including \textit{flask}, \textit{pandas}, \textit{pytorch}, \textit{scikit}, and \textit{seaborn}—into coding questions. When prompting LLMs to generate challenging yet clear and self-contained questions, we implemented a quality control mechanism allowing the model to abstain when the provided documentation proves insufficient for crafting high-quality coding questions. The complete prompt template is available in Appendix \ref{appendix: docs template}.
\color{black}

\textbf{More Questions.} We further expand coding questions by employing \textsc{Magpie} \cite{xu2024magpie} using seven open-source LLMs. We employ LLM annotators to classify generated questions, retaining only high-quality examples under "Algorithm Implementation" or "Function Generation" categories. We name this subset \textbf{Filter}. Details of this subset can be found in Appendix \ref{appendix: filter subset details}. In addition, we synthesize more questions from existing \textit{Package Instruct} \cite{Huang2024OpenCoderTO} and \textit{Evol Instruct} \cite{luo2023wizardcoder} synthetic datasets, and create two subsets named as \textbf{Package} and \textbf{Evol}.

\textbf{Deduplication.} After generating questions, we perform semantic deduplication within each subset by utilizing the \texttt{all-mpnet-base-v2} embedding model to project all questions into an embedding space. We then compute nearest-neighbor distances using FAISS \citep{douze2024faiss}, and filter out questions that surpass a predefined similarity threshold with existing entries.

\subsection{Step 2: Solution \& Test Generation}

To generate verifiably correct coding solutions and unit tests $(sol, test)$ for questions from Step 1, we employ a \textbf{self-verification} procedure as detailed below. To ensure quality of solution and tests, we first employ \textit{GPT-4o-0513} (which achieves state-of-the-art performance among non-reasoning models on the BigCodeBench Leaderboard \citep{zhuo2024bigcodebench}) to generate both solution and test, then execute these unit tests to validate the correctness of the solution. Additionally, we perform branch coverage analysis using the \textsc{pytest-cov} framework to ensure the diversity of test cases. Only question-solution-test triplets that pass self-verification and achieve 100\% branch coverage are retained. The prompt template is provided in Appendix \ref{appendix: solution template}.

Since our goal is to generate verifiable training data that is challenging and diverse in coverage, a key challenge arises: even state-of-the-art models cannot guarantee bug-free code or ensure that solutions pass their unit tests. Simply discarding questions that fail self-verification risks eliminating many challenging ones, potentially biasing our dataset towards a distribution of simple problems.

Our solution to address this challenge is to assign additional self-verification attempts for hard questions. For each question from Step 1, we allow up to a maximum of $n$ attempts (where $n=10$ in our experiments) to generate a solution that passes its unit tests. 
Importantly, each attempt regenerates both the solution and its corresponding unit tests from scratch, since if the initial tests contain errors, all subsequent solutions would fail regardless of their correctness. To maintain the quality of regenerated test cases, we only retain new $(sol, test)$ that achieve equal or higher branch coverage compared to previous attempts. This ensures that regenerated tests do not become progressively simpler or less comprehensive.

We note that our approach can preserve challenging questions while naturally assigning difficulty labels based on the success rate across attempts. Questions that fail to generate correct solutions after $n$ attempts are discarded as they likely contain inherent flaws. Upon completing Step 2, we obtain a collection of 279K verified triplets.

    
\subsection{Step 3: Post-training Data Synthesis}




In creating LLM post-training data for coding tasks, there is a gap between coding questions (e.g., LeetCode) and training data. While coding questions are primarily expressed in natural language, training data needs to accommodate non-natural-language formats such as function calls and tool interactions. To address this disparity, we propose an LLM-based style converter to enhance the diversity of question formats. Specifically, we reformat each question $q$ by taking its solution and test as inputs, $q'=\text{LLM}(q, sol, test)$, structuring them as Python completion tasks with function signatures and examples. Each reformatted question is paired with its original solution and test cases to form new triplets $(q', sol, test)$. This process results in the creation of 168K additional triplets, increasing our total to 447K, which are readily available for RL training.



Motivated by recent advances in reasoning models \citep{sky_t1_2025, bespoke_32b}, we further generate an SFT dataset for post-training by leveraging the questions in these triplets, using DeepSeek R1 \citep{deepseekr1} as the response generator to generate Chain-of-Thought responses. To ensure the quality of the generated responses, we generate 3 times for each question and perform test-based reject sampling, yielding a large-scale, high-quality, and verifiably correct SFT dataset for coding. We refer to this dataset as \datanamen-SFT.
\section{Analysis} \label{sec: dataset stats}


In what follows, we conduct a comprehensive analysis to demonstrate the effectiveness of \dataname{} in generating diverse, challenging, and correct question-solution-test triplets. 

\subsection{Pipeline Analysis}
\label{sec: pipeline analysis}

\textbf{Effectiveness of Self-Verification.} To evaluate the reliability of our self-verification pipeline, we conduct experiments on the MBPP validation dataset \citep{austin2021program} and the LiveCodeBench-V5 test set \citep{jain2024livecodebench}. For MBPP, we utilize all 90 coding questions that include ground-truth unit tests manually written and verified by humans. For LiveCodeBench-V5, we select all 381 questions that provide starter code and functional test cases.

Following our pipeline, we generate both solutions and unit tests for these questions. After applying our Self-Verification approach, 80 solutions (88.9\%) pass self-verification for MBPP, while 190 solutions (49.9\%) pass self-verification for LiveCodeBench-V5. We then evaluate these retained solutions against the ground-truth unit tests.

We note that due to inherent ambiguity in MBPP questions, some solutions that follow correct logic fail assertions because of mismatched input formats or numerical precision differences. After manual review of these edge cases, 78 out of 80 solutions pass all ground-truth unit tests, achieving a 97.5\% pass rate for MBPP. For LiveCodeBench-V5, 189 solutions (99.47\%) successfully pass all tests. This high success rate demonstrates the effectiveness of our pipeline in generating verified solutions. Please refer to Appendix \ref{sec: failed mbpp test} for detailed analysis of the failure cases.

\textbf{Effectiveness of Allocating Additional Attempts to Challenging Questions.} To assess the impact of allocating more attempts on solution \& test generation, we adopt the $Pass@k$ metric commonly used in model evaluation literature \cite{austin2021program, chen2021evaluating}. Specifically, we measure the proportion of questions for which at least one out of $k$ solutions successfully passes its self-verification in Step 2. 

\begin{figure}[htpb]
    \centering
    \includegraphics[width=1.0\linewidth]{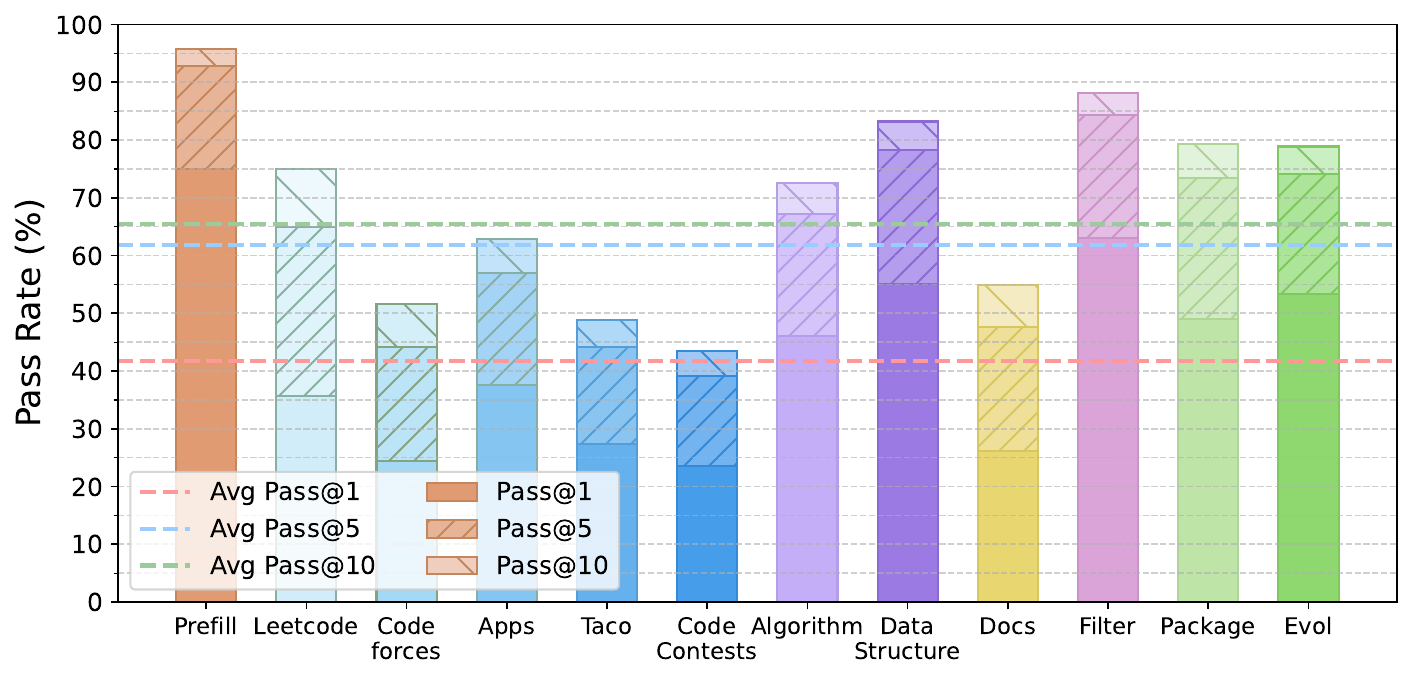}
    \caption{Statistics on pass rates via self-verification in Step 2 by subset with varying number of attempts.}
    \label{fig:pass_rate}
    \vspace{-1em}
\end{figure}

The experimental results are illustrated in Figure \ref{fig:pass_rate}, where we report $Pass@k$ by subsets and on average. We observe that while $Pass@1$ yields a low pass rate, increasing the number of trials from 1 to 5 results in an average pass rate increase of over 20\%, and further increasing to 10 trials boosts the pass rate by an additional 4\%. Notably, for more challenging tasks, such as Codeforces and Docs subsets, increasing the number of attempts significantly enhances pass rates. In contrast, simpler tasks like those in the Prefill subset show more modest gains from additional attempts. We emphasize that this scaling in attempts enables \dataname{} to retain more challenging questions that would otherwise be discarded.

\begin{figure}[htpb]
\centering
\includegraphics[width=1.0\linewidth]{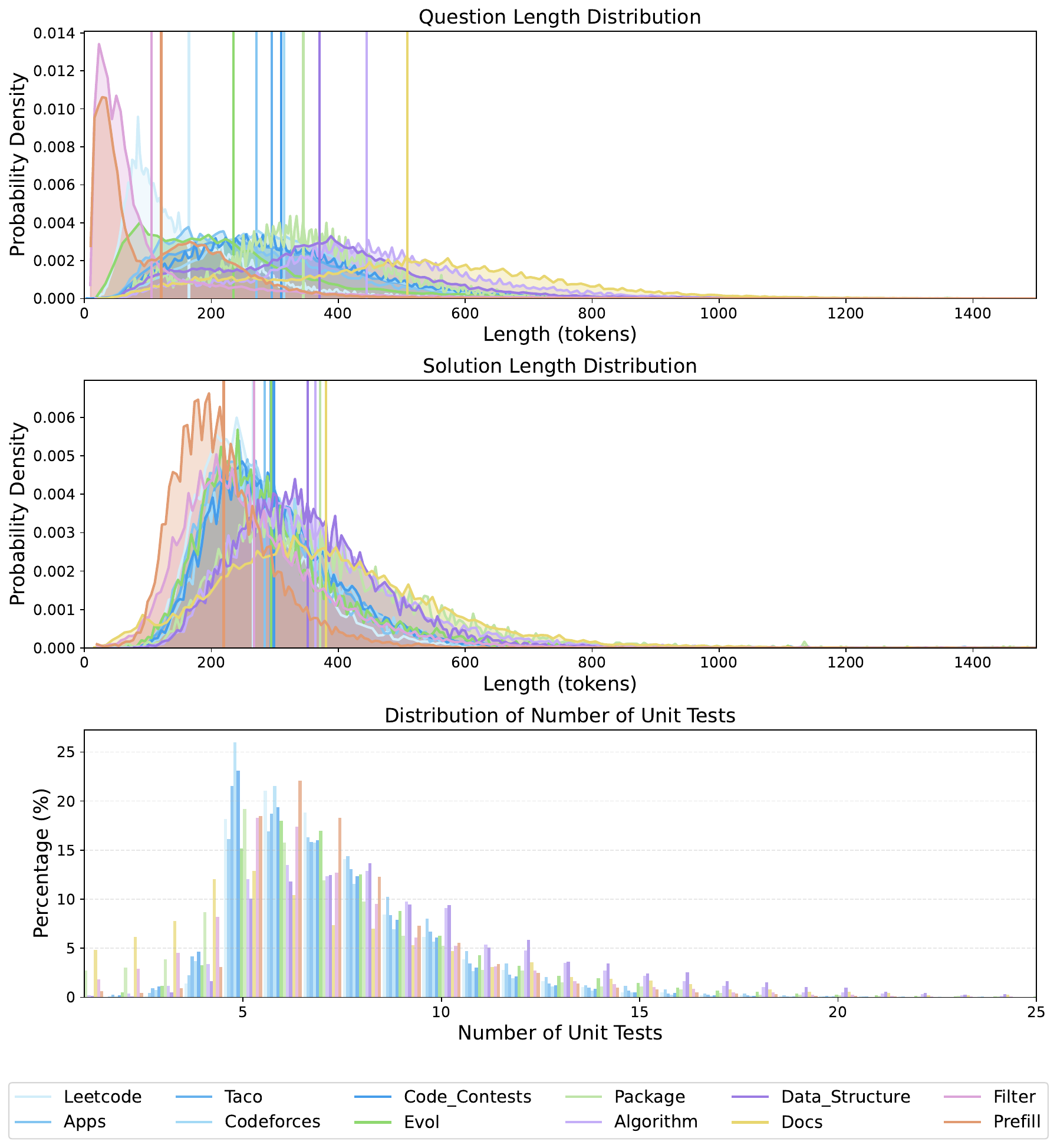}
\caption{Distribution of token lengths for questions and responses, along with unit test counts across subsets.}
\label{fig:length_distribution}
\vspace{-1em}
\end{figure}

\subsection{Dataset Statistics and Analysis}

In what follows, we analyze \dataname~from the perspective of token length, diversity, difficulty, potential contamination, and data-flow analysis.

\textbf{Token Length and Unit Test Statistics.} Figure \ref{fig:length_distribution} presents the distribution of token counts for questions and solutions across different subsets. We also provide unit test statistics of \datanamen. Each coding question in \dataname~contains an average of 7.52 unit tests.

\textbf{Diversity.} We analyze the diversity of \datanamen-V1's question distribution by comparing it with four baseline datasets: OSS Instruct \cite{wei2024magicoder}, ACECoder \cite{AceCoder}, Educational Instruct \cite{Huang2024OpenCoderTO}, and Package Instruct \cite{Huang2024OpenCoderTO}. Using the \texttt{all-mpnet-base-v2} embedding model\footnote{\url{https://huggingface.co/sentence-transformers/all-mpnet-base-v2}}, we encode the questions and visualize their distribution using t-SNE \citep{van2008visualizing} to create a two-dimensional representation.

\begin{figure}[htpb]
    \centering
    \includegraphics[width=1.02\linewidth]{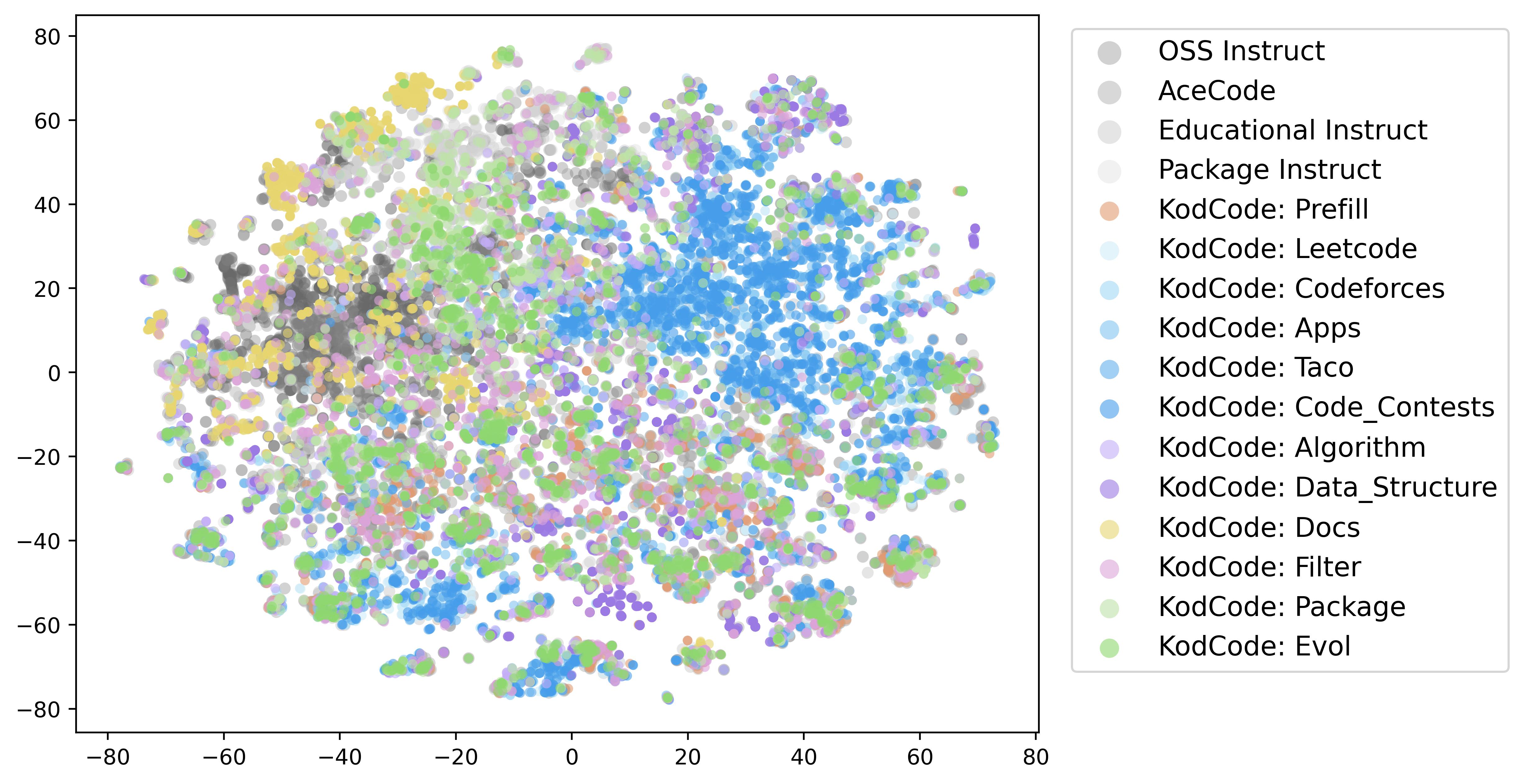}
    \caption{Comparison of t-SNE visualization between \dataname{} (by subset) and baseline datasets (OSS Instruct, ACECoder, Educational Instruct, and Package Instruct), with 2,000 sampled instructions per dataset.}
    \label{fig:tsne}
\end{figure}

The visualization in Figure \ref{fig:tsne} reveals two key observations. First, \datanamen's question distribution (shown in color) spans the entire space, while baseline datasets (in gray) cluster primarily in the upper left region, demonstrating \datanamen's broader topical diversity. Second, the Algorithm and Filter subsets of \datanamen~show comprehensive coverage across the entire space, validating their role in enhancing \datanamen's overall diversity.

\begin{figure}[htpb]
\centering
\includegraphics[width=1.02\linewidth]{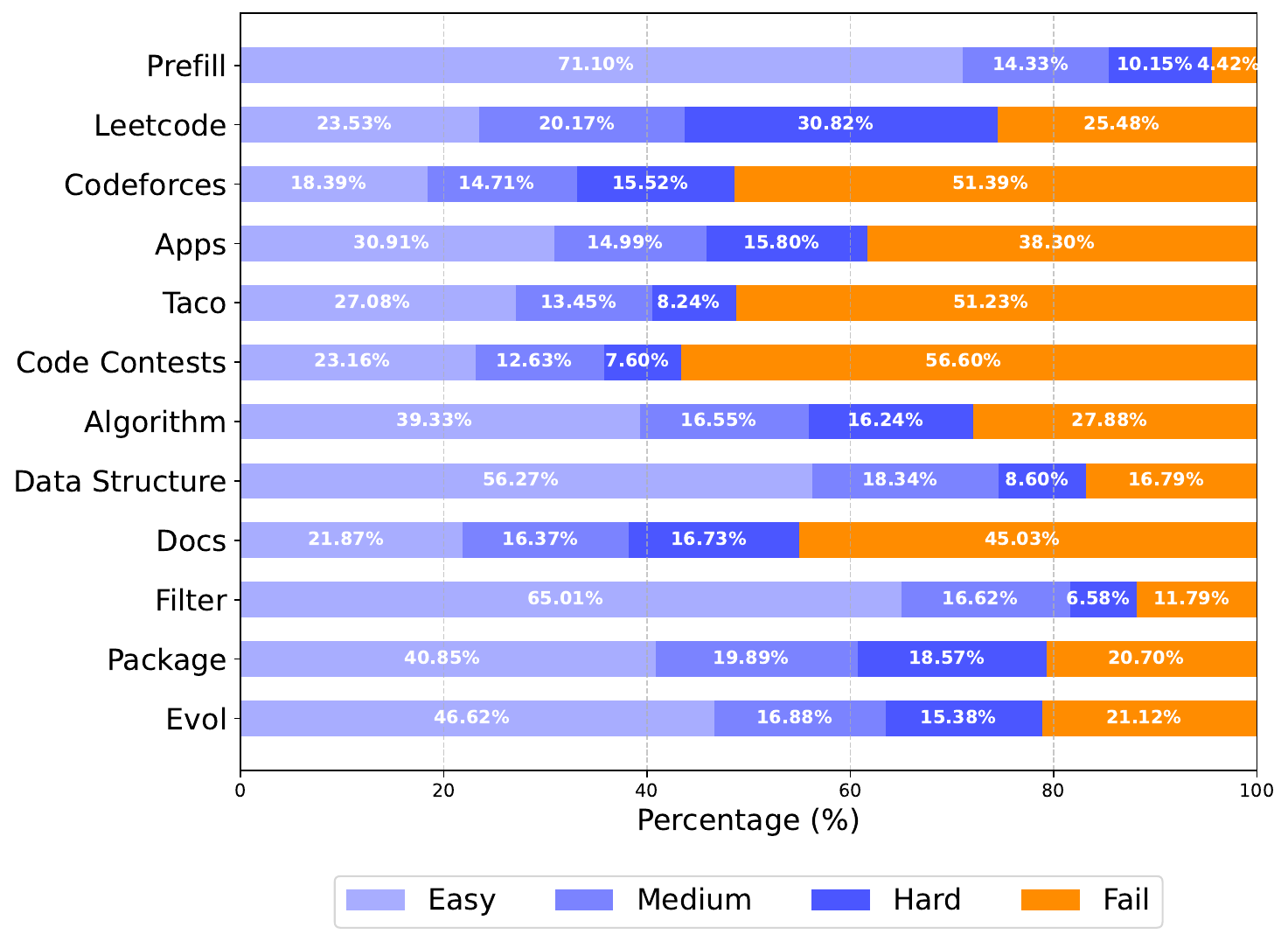}
\caption{Difficulty distribution across subsets measured by pass rates.}
\label{fig:difficulty}
\vspace{-1em}
\end{figure}

\textbf{Difficulty.} We analyze the difficulty distribution across \dataname{} subsets by examining the success rate of $(sol, test)$ pairs in self-verification across $n=10$ attempts per question, as shown in Figure \ref{fig:difficulty}. We categorize questions into four difficulty levels: easy (pass rate >2/3), medium (1/3 to 2/3), hard (<1/3), and fail (all failures). The analysis reveals that the Prefill subset are typically the easiest, while Codeforces, Taco, and Code Contests subsets are more challenging, as shown in their higher failure rates and larger proportions of hard questions.

\textbf{Contamination Analysis.} We evaluate potential contamination between \dataname{} and existing evaluation benchmarks, including HumanEval \citep{chen2021evaluating}, MBPP \citep{austin2021program}, BigCodeBench \citep{zhuo2024bigcodebench}, and LiveCodeBench (V5) \cite{jain2024livecodebench}. For each question in \dataname{}, we identify the most similar benchmark question using cosine similarity of embeddings, considering a question contaminated if the similarity exceeds 0.95.

\begin{figure}[htpb]
    \centering
    \includegraphics[width=1.0\linewidth]{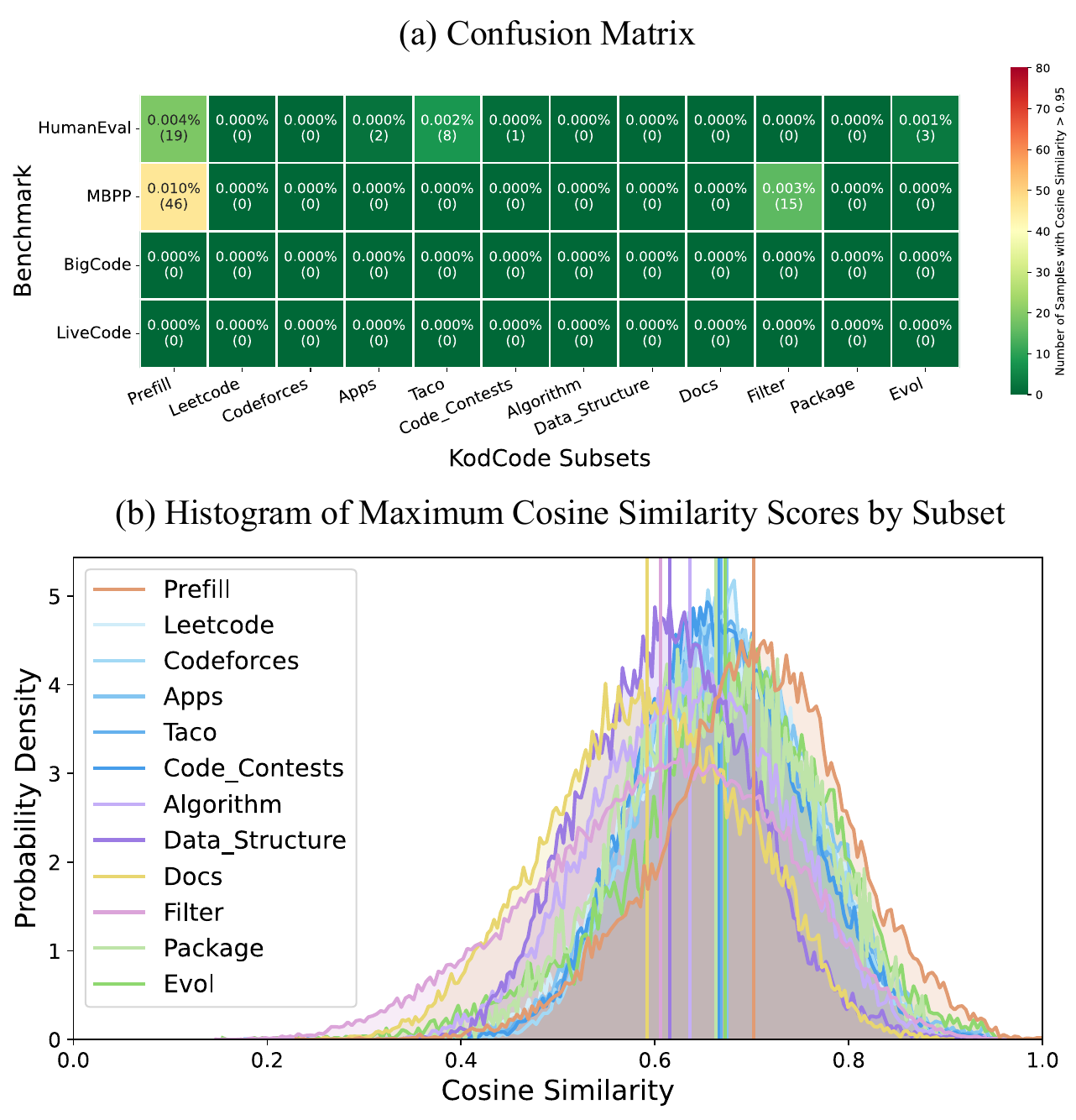}
    \caption{Contamination analysis between \dataname{} subsets and existing benchmarks. (a) Confusion matrix showing the percentage and absolute number (in parentheses) of contaminated samples with cosine similarity > 0.95. (b) Distribution of maximum cosine similarity scores across different \dataname subsets, with horizontal lines indicating subset averages.}
    \label{fig:contamination}
    \vspace{-1em}
\end{figure}

As shown in Figure \ref{fig:contamination}, our analysis reveals three findings. First, the contamination rate is minimal, with only 94 potentially contaminated questions out of 447K. Second, most overlaps occur between the Prefill subset and simple Python questions from HumanEval/MBPP. Third, the histogram in Figure \ref{fig:contamination}-b shows that Prefill has the highest average maximum cosine similarity, while Docs has the lowest. We provide examples of contaminated questions in Appendix \ref{appendix: contamination examples} and exclude these cases from our performance evaluation in Section \ref{sec: experiments}.

\begin{figure}[htpb]
\centering
\includegraphics[width=1.0\linewidth]{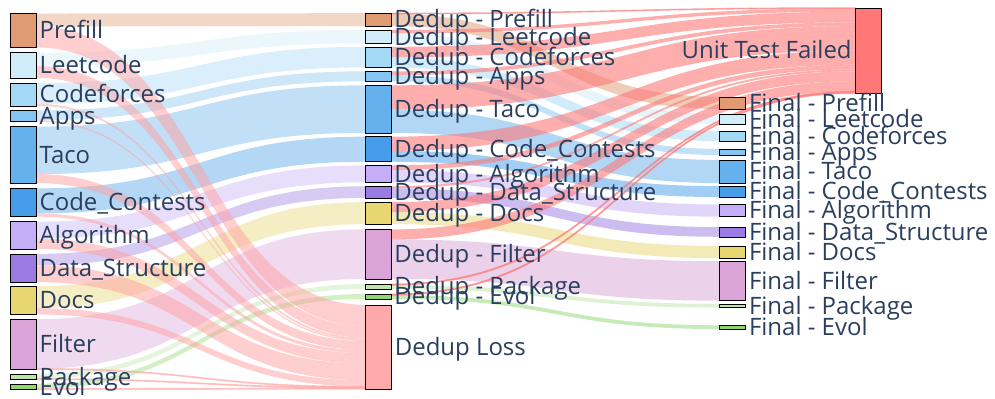}
\caption{Data flow visualization through our pipeline: from initial subsets (left) through deduplication (middle) to final filtered sets after reject sampling (right). Red paths indicate discarded instances.}
\label{fig:survival rate}
\end{figure}

\textbf{Data Flow Analysis.} Figure \ref{fig:survival rate} presents a Sankey diagram illustrating data flow throughout our synthesis dataset generation pipeline. Over 25\% of instances are eliminated during Step 1's deduplication process, with the Prefill subset showing the highest redundancy rate (over 50\% discarded). In Step 2, instances failing unit tests are filtered out, with higher difficulty subsets (such as Codeforces and Taco) showing higher rejection rates.

\begin{table*}[htpb]
\centering
\renewcommand{\arraystretch}{1.2}
\resizebox{\textwidth}{!}{
\begin{tabular}{cccccccccccccc}
\toprule
\multicolumn{2}{c}{\multirow{2}{*}{\textbf{Model Name}}} & \multicolumn{2}{c}{\textbf{HumanEval}} & \multicolumn{2}{c}{\textbf{MBPP}} & \multicolumn{2}{c}{\textbf{BigCodeBench-C}} & \multicolumn{2}{c}{\textbf{BigCodeBench-I}} & \multicolumn{3}{c}{\textbf{LiveCodeBench (v5)}} & \multirow{2}{*}{\textbf{Average}} \\
\multicolumn{2}{c}{} & Base & Plus & Base & Plus & Full & Hard & Full & Hard & Easy & Medium & Hard & \\
\hline
\multirow{5}{*}{\begin{tabular}[c]{@{}c@{}}\textbf{Non-Reasoning} \\ \textbf{Models}\end{tabular}} 
& Llama-3.1-Tulu-3-70B & 83.5 & 78.0 & 75.9 & 65.9 & 55.0 & 25.0 & 43.4 & 20.9 & 61.7 & 15.4 & 3.7 & 50.15 \\
& Llama-3.3-70B-Instruct & 82.9 & 77.4 & 87.3 & 73.0 & 57.9 & 29.1 & 47.0 & 26.4 & 81.4 & 21.1 & 8.5 & 55.5 \\
& Qwen2.5-32B-Instruct & 89.6 & 79.9 & 87.8 & 73.8 & 53.2 & 26.4 & 45.4 & 22.3 & 80.3 & 35.9 & 8.5 & 56.27 \\
& Qwen2.5-72B-Instruct & 87.8 & 81.1 & 90.2 & 76.2 & 57.5 & 33.1 & 46.1 & 21.6 & 69.9 & 39.9 & 7.0 & 57.15 \\
& Qwen2.5-Coder-32B-Instruct & 90.9 & 85.4 & \textbf{90.2} & \textbf{77.0} & 57.6 & 31.1 & 49.4 & 25.7 & 80.6 & 39.0 & 8.2 & 59.25 \\
\hline
\multirow{5}{*}{\begin{tabular}[c]{@{}c@{}}\textbf{Reasoning} \\ \textbf{Models}\end{tabular}} 
& Sky-T1-32B-Preview & 87.8 & 80.5 & 87.3 & 74.1 & 51.0 & 27.0 & 44.2 & 24.3 & 81.4 & 21.1 & 8.5 & 54.9 \\
& QwQ-32B-Preview & 87.8 & 82.3 & 84.4 & 69.8 & 53.9 & 26.4 & 38.8 & 23.0 & 90.0 & 51.7 & 10.0 & 56.75 \\
& DeepSeek-R1-Distill-Qwen-32B & 87.8 & 81.1 & 83.9 & 69.3 & 55.1 & 27.7 & 44.1 & 23.0 & 83.9 & 52.9 & 17.8 & 57.51 \\
& DeepSeek-R1-Distill-Llama-70B & 89.0 & 80.5 & 81.7 & 68.8 & 53.5 & 25.7 & 43.9 & 25.7 & 88.5 & \textbf{56.2} & \textbf{18.9} & 57.79 \\
& Bespoke-Stratos-32B & 88.4 & 83.5 & 88.1 & 75.1 & 56.2 & 33.1 & 47.3 & 27.0 & 86.7 & 49.5 & 10.4 & 59.64 \\
\hline
\multirow{2}{*}{\begin{tabular}[c]{@{}c@{}}\textbf{\dataname{}}\end{tabular}} 
& \datanamen-32B-SFT-50K & \textbf{92.7} & 85.4 & 89.9 & 76.2 & \textbf{59.8} & \textbf{37.8} & \textbf{51.1} & \textbf{32.4} & 87.8 & 35.9 & 6.7 & 61.22 \\
& \datanamen-32B-SFT-Hard-18K & 90.9 & \textbf{86.6} & 89.2 & \textbf{77.0} & 59.7 & 37.2 & 50.5 & 31.1 & \textbf{90.7} & 39.3 & 5.6 & \textbf{61.26} \\ 
\bottomrule
\end{tabular}}
\caption{This table compares the model performances between \datanamen-tuned models and strong baseline models across various benchmarks. Average scores are computed by first averaging sub-metrics within each benchmark and then taking the mean across all five benchmarks (HumanEval, MBPP, BCB-Complete, BCB-Instruct, and LiveCodeBench). Bold numbers indicate the best performance for each metric. \datanamen-tuned models outperform larger baseline models, demonstrating the dataset's high quality and diversity.}
\label{tab:model_performance}
\end{table*}

\section{Performance Evaluation}
\label{sec: experiments}

To evaluate the performance of the \dataname{} dataset, we conduct both supervised fine-tuning using Qwen2.5-Coder-32B-Instruct \cite{hui2024qwen25coder} on the \datanamen-SFT dataset, and GRPO \cite{shao2024deepseekmath} using Qwen2.5-7B-Instruct-1M \cite{qwen2.5-1m} and Qwen2.5-Coder-7B-Instruct \cite{hui2024qwen25coder} on the \datanamen{} dataset.
We compare the model's performance across several widely used code generation benchmarks against existing model baselines.

\subsection{Experimental Setup}

\textbf{SFT Setup.} We evaluate two variants of \datanamen-SFT. For preprocessing, we exclude R1 responses that are too long, too short, or implement class-based solutions rather than functional implementations. The first version, \datanamen-SFT-50K, contains 50K instruction-response pairs selected based on the empirical results of mixing data from different sources and difficulties. The second version, \datanamen-SFT-Hard-18K, contains all 18K instructions that are labeled as hard after applying the same preprocessing filters. We refer to the models fine-tuned on these two datasets as \datanamen-32B-SFT-50K, and \datanamen-32B-SFT-Hard-18K. We use a cosine learning rate schedule with a maximum learning rate of $1 \times 10^{-5}$ when fine-tuning the Qwen-2.5 model.
The maximum sequence length is 16384. The detailed training configurations can be found in Appendix \ref{appendix: Instruction-Tuning Setups}.

\textbf{RL Setup.} We conduct RL experiments on both Qwen2.5-7B-Instruct-1M \cite{qwen2.5-1m} and Qwen2.5-Coder-7B-Instruct \cite{hui2024qwen25coder} using 9.5K randomly selected samples from \datanamen as the training set, and 0.5K as the validation set. We refer to this dataset as \dataname-RL-10K. We perform GRPO \cite{shao2024deepseekmath} for 256 steps with actor learning rate of $5 \times 10^{-7}$, 16 rollouts per question, a batch size of 256, max response length of 4096, and apply KL coefficient of 0.001. We assign binary rewards for RL: 1 if the model's solution passes all unit tests, and 0 otherwise.

\textbf{Baselines.} We compare our \datanamen-tuned model against several strong baselines, including non-reasoning models (Llama-3.3-70B-Instruct 
\citep{dubey2024llama}, Llama-3.1-Tulu-3-70B \citep{lambert2025tulu3pushingfrontiers}, Qwen-2.5-32B/72B-Instruct \citep{qwen2.5}, Qwen-2.5-Coder-32B-Instruct \citep{hui2024qwen25coder}), and reasoning models (QwQ-32B-Preview \citep{qwq-32b-preview}, DeepSeek-R1-Distill-Llama-70B \citep{deepseekr1}, Sky-T1-32B-Preview \citep{sky_t1_2025}, and Bespoke-Stratos-32B \citep{bespoke_32b}).

\textbf{Benchmarks and Evaluation Setups.} We evaluate models on HumanEval(+) \citep{chen2021evaluating, liu2024your}, MBPP(+) \citep{austin2021program, liu2024your}, BigCodeBench \citep{zhuo2024bigcodebench}, and LiveCodeBench (V5) \citep{jain2024livecodebench}, each designed to assess different aspects of code generation, including functional correctness, external library usage, and competitive programming challenges. We use EvalPlus for HumanEval(+) and MBPP(+) evaluation, and Skythought-Evals \citep{li2025llmseasilylearnreason} for LiveCodeBench evaluation. We evaluate performance on both Complete and Instruct subsets of BigCodeBench.

We follow the official setups in each benchmark and evaluate all models using greedy decoding with a maximum generation length of 16,384 tokens. We follow the official chat templates of the model during inference.





\subsection{Experimental Results}


\paragraph{Model fine-tuned using \datanamen-SFT outperforms baseline models.} Table \ref{tab:model_performance} presents a comparative analysis of Qwen-2.5-32B-Coder-Instruct fine-tuned on \datanamen-SFT against various baseline models. Our fine-tuned SFT models consistently outperform all baselines, including larger models, across both the Complete and Instruct subsets of BigCodeBench. Specifically, for BigCodeBench-C, our model achieves 59.8\% (+1.9\%) in the Full category and 37.8\% (+4.7\%) in the Hard category, compared to the strongest baseline. For BigCodeBench-I, it achieves 51.1\% (+1.7\%) in the Full category and 32.4\% (+5.4\%) in the Hard category, demonstrating strong performance across different evaluation settings. Additionally, on HumanEval, our model reaches 92.7\%, surpassing Qwen2.5-Coder-32B-Instruct (90.9\%) by 1.8\%. For LiveCodeBench-Easy, our model attains 90.7\%, exceeding all baselines, including QwQ-32B-Preview (90.0\%). Overall, our model achieves an average score of 61.26\%, the highest among all evaluated models, highlighting the quality and diversity of our \dataname dataset and its effectiveness in improving code generation performance across various benchmarks.


\paragraph{Effectiveness of hard coding questions.}
 To validate the impact of challenging instances in \datanamen, we compare models trained on two different 10K sample sets: one randomly sampled from the \datanamen-SFT-50K dataset (named as \datanamen-SFT-10K) and another from the \datanamen-SFT-Hard-18K dataset (named as \datanamen-SFT-Hard-10K). As shown in Table \ref{tab: 10K_ablation}, \datanamen-SFT-Hard-10K outperforms \datanamen-SFT-10K on BigCodeBench-I Hard (31.8\% vs. 27.7\%, +4.1\%) and BigCodeBench-I Full (50.6\% vs. 49.9\%, +0.7\%), confirming that exposure to difficult coding problems improves model robustness. Similarly, for BigCodeBench-C Hard, \datanamen-SFT-Hard-10K achieves 39.9\%, surpassing \datanamen-SFT-10K (38.5\%, +1.4\%). On LiveCodeBench Hard, \datanamen-SFT-Hard-10K leads (6.3\% vs. 4.8\%, +1.5\%), reinforcing that hard samples do enhance performance on complex programming tasks.


\begin{table}[t!]
\centering
\vspace{0.5em}
\resizebox{0.5\textwidth}{!}{
\begin{tabular}{lccc}
\toprule
\textbf{Benchmarks} & \textbf{\shortstack{\datanamen\\SFT-Hard-10K}} & \textbf{\shortstack{\datanamen\\SFT-10K}} & \textbf{\shortstack{\datanamen-SFT\\NoConvert-10K}} \\
\midrule
BigCodeBench-C Full   & 60.4& \textbf{61.1}& 60.3 \\
BigCodeBench-C Hard   & \textbf{39.9}& 38.5& 35.1 \\
BigCodeBench-I Full   & \textbf{50.6}& 49.9& 49.6 \\
BigCodeBench-I Hard   & \textbf{31.8}& 27.7& 28.4 \\
\midrule
LiveCodeBench Easy         & \textbf{87.8}& \textbf{87.8}& 86.4 \\
LiveCodeBench Medium       & \textbf{35.3}& 32.6& 32.6 \\
LiveCodeBench Hard         & \textbf{6.3}& 4.8& 5.6 \\
\bottomrule
\end{tabular}
}
\caption{Ablation study of data selection when fine-tuning Qwen2.5-Coder-32B-Instruct on \datanamen-SFT. Each model is trained on 10K sampled data. Bold numbers indicate best performance for each metric.}
\label{tab: 10K_ablation}
\end{table}


\paragraph{Effectiveness of style converter in \datanamen~generation process.}
To assess the impact of the style converter, we remove all instances processed by the style converter from the \datanamen-SFT-50K dataset and randomly sample 10K instances from the remaining datasets for fine-tuning, naming the resulting dataset \datanamen-SFT-NoConvert-10K. As shown in Table \ref{tab: 10K_ablation}, \datanamen-SFT-NoConvert-10K performs lower on BigCodeBench-C Full (60.3\% vs. 61.1\%) and BigCodeBench-C Hard (35.1\% vs. 38.5\%), indicating that removing style variations slightly reduces performance. In LiveCodeBench Easy, \datanamen-SFT-NoConvert-10K scores 86.4\%, lower than \datanamen-SFT-10K (87.8\%). This result is consistent with the findings of \citep{li2025llmseasilylearnreason}, highlighting that the question format plays a role in the performance of the LLM code.

\paragraph{Effectiveness of \datanamen{} on RL.}

\begin{table*}[htpb]
\centering
\vspace{1em}
\renewcommand{\arraystretch}{1.2}
\resizebox{\textwidth}{!}{
\begin{tabular}{ccccccccccccc}
\toprule
\multirow{2}{*}{\textbf{Model}} & \multicolumn{3}{c}{\textbf{LiveCodeBench}} & \multicolumn{2}{c}{\textbf{BCB-Complete}} & \multicolumn{2}{c}{\textbf{BCB-Instruct}} & \multicolumn{2}{c}{\textbf{HumanEval}} & \multicolumn{2}{c}{\textbf{MBPP}} & \multirow{2}{*}{\textbf{Average}} \\
\cmidrule(lr){2-4} \cmidrule(lr){5-6} \cmidrule(lr){7-8} \cmidrule(lr){9-10} \cmidrule(lr){11-12}
& Easy & Medium & Hard & Full & Hard & Full & Hard & Base & Plus & Base & Plus & \\
\hline
Qwen2.5-Coder-7B-Instruct & 57.4 & 23.0 & 4.4 & 52.0 & 21.6 & 41.8 & 19.6 & 91.5 & 85.4 & 83.1 & 71.7 & 52.32 \\
+ RL KodCode-10K (Step 128) & 65.2 & 21.1 & 4.1 & 52.5 & 25.7 & 42.2 & 20.3 & 90.9 & 86.0 & 84.9 & 72.8 & 53.56 \\
+ RL KodCode-10K (Step 256) & 64.5 & 19.9 & 3.3 & 53.7 & 27.0 & 42.9 & 21.6 & 90.2 & 85.4 & 86.5 & 74.1 & \textbf{53.99} \\
\midrule
Qwen2.5-7B-Instruct-1M & 57.7 & 12.4 & 3.7 & 45.3 & 14.2 & 36.6 & 17.6 & 86.0 & 79.3 & 78.8 & 69.3 & 47.63 \\
+ RL KodCode-10K (Step 128) & 60.2 & 19.0 & 2.6 & 47.0 & 19.6 & 36.7 & 13.5 & 90.2 & 83.5 & 81.0 & 70.9 & 49.69 \\
+ RL KodCode-10K (Step 256) & 57.0 & 18.7 & 3.0 & 48.2 & 19.6 & 36.8 & 12.8 & 91.5 & 86.0 & 82.8 & 72.8 & \textbf{50.30} \\
\bottomrule
\end{tabular}}
\caption{This table evaluates the performance of models before and after training with GRPO using \datanamen-RL-10K. Notable performance gains are observed across most benchmarks compared to the baselines.}
\label{tab:model_performance_rl}
\end{table*}

Table \ref{tab:model_performance_rl} compares the performance of models before and after GRPO using \datanamen-RL-10K. We observe significant performance improvement on most of the benchmarks compared to the original model after RL. In addition, we observe that continuing to increase the training steps can further enhance the model performance. This indicates the effectiveness of \datanamen{} for RL training.
\section{Related Work}

\paragraph{Synthetic Data Generation for Code LLMs.}
High-quality training data is crucial for LLM post-training \citep{zhou2023lima, alpaca}. Given the time and resource cost of human data collection \cite{Dolly}, synthetic data generation has emerged as a promising alternative. This approach leverages LLMs to produce synthetic instructions by expanding a small set of human-annotated seed instructions through few-shot prompting \citep{wang-etal-2023-self-instruct, alpaca, xu2023wizardlm, xu-etal-2023-baize, wang2024codeclm, sun2024principle}. While synthetic data generation has been widely explored for alignment \cite{xu2024magpie, xu2023wizardlm, ding2023ultrachat, cui2023ultrafeedback} and mathematics \cite{numina_math_datasets, yue2024mammoth2scalinginstructionsweb, toshniwal2024openmathinstruct2acceleratingaimath} with millions of instances available, high-quality synthetic coding datasets remain scarce. Recently, several open-source coding datasets have been proposed by the community, including Code Alpaca \cite{codealpaca}, OSS Instruct \cite{wei2024magicoder}, Evol Instruct \cite{luo2023wizardcoder}, and Package Instruct \cite{Huang2024OpenCoderTO}. However, these resources are still limited in terms of diversity, difficulty, and scale. We present a comprehensive comparison between \datanamen~and existing open-source coding datasets in Table \ref{tab:compare with baselines}.

\paragraph{Code Generation with Execution Feedback.} Ensuring the correctness of code generated by LLMs remains a critical challenge. \citet{yang2024intercode} explores execution feedback within Docker environments, enabling models to rectify syntactic and logical errors by iteratively refining their outputs based on runtime execution results. \citet{zheng2025opencodeinterpreterintegratingcodegeneration} integrates code generation with execution and refinement to improve the quality of generated code.

\paragraph{LLM-based Unit Test Generation.} While execution feedback helps identify code issues, unit tests play a complementary role by proactively assessing code correctness. Yang et al. \citep{yang2024evaluationlargelanguagemodels} provide a comprehensive empirical analysis on LLMs' capabilities in unit test generation. EvalPlus \cite{liu2024your} enhances code evaluation by combining LLM-generated test cases with mutation-based expansion, increasing test diversity and rigor. Huang et al. \citep{huang2023enhancing} propose a multi-perspective self-consistency framework to select optimal code solutions based on self-generated tests. Recently, Wei et al. \citep{wei2024selfcodealign} introduced an approach to align a base code model by generating solutions and verifying their correctness through self-generated unit tests. OpenCoder \citep{Huang2024OpenCoderTO} employs a teacher model to generate multiple test cases for each code snippet, executing them in a Python interpreter and filtering out failing samples to ensure reliability. 
Jiao et al. \citep{jiao2025preferenceoptimizationreasoningpseudo} enhance the reasoning capabilities of LLMs by evaluating solutions to reasoning problems via LLM-generated test case.
AceCoder \citep{AceCoder} prompts GPT-4o-mini to generate unit tests, with Qwen 2.5 Coder-32B Instruct acting as a verifier to eliminate incorrect test cases.

\section{Conclusion and Future Work}

In this work, we presented \datanamen, a large-scale synthetic dataset of 447K diverse coding questions paired with verified solutions and unit tests. Our three-step synthesis pipeline—comprising coding question generation, solution and test case refinement via self-verification, and post-training data synthesis—ensures both the diversity and quality of training data for high-performing coding language models. 
Through detailed analysis, we validate our pipeline's effectiveness and thoroughly examine the dataset's attributes.
Comprehensive experiments demonstrate that models fine-tuned on \datanamen~not only achieve state-of-the-art performance across multiple benchmarks (HumanEval(+), MBPP(+), BigCodeBench, and LiveCodeBench) but also outperform larger models.

Future work will focus on three directions. First, we plan to scale up the dataset with more challenging problems, as our findings indicate that difficult instances significantly improve model performance. Second, we aim to investigate optimal strategies for post-training data selection. Finally, we will explore methods for generating repository-level synthetic data to further enhance coding LLMs. 

\section*{Acknowledgment}

This work is partially supported by the Office of Naval Research (ONR) under grant N0014-23-1-2386, the Air Force Office of Scientific Research (AFOSR) under grant FA9550-23-1-0208, and the National Science Foundation (NSF) AI Institute for Agent-based Cyber Threat Intelligence and Operation (ACTION) under grant IIS 2229876.

This work is supported in part by funds provided by the National Science Foundation, Department of Homeland Security, and IBM. 
Any opinions, findings, and conclusions or recommendations expressed in this material are those of the author(s) and do not necessarily reflect the views of the NSF or its federal agency and industry partners.
\section*{Limitations}

While models fine-tuned on \dataname achieve state-of-the-art performance across most coding benchmarks, their performance on LiveCodeBench-Hard remains limited. This gap likely stems from insufficient representation of highly challenging competition-level programming problems in our dataset, which are prevalent in LiveCodeBench-Hard. Future work could explore methods to synthesis highly challenging coding problems.

\section*{Ethical Statement}

Our \dataname dataset enhances code generation of code LLMs through diverse and challenging instruction-solution-test triplets. We do not introduce or endorse any applications that could cause harm or be misused.
This paper does not present any ethical concerns.



\bibliography{custom}

\clearpage


\appendix
\section{Additional Information of the Filter Subset.}
\label{appendix: filter subset details}

\begin{figure}[htpb]
    \centering
    \includegraphics[width=0.9\linewidth]{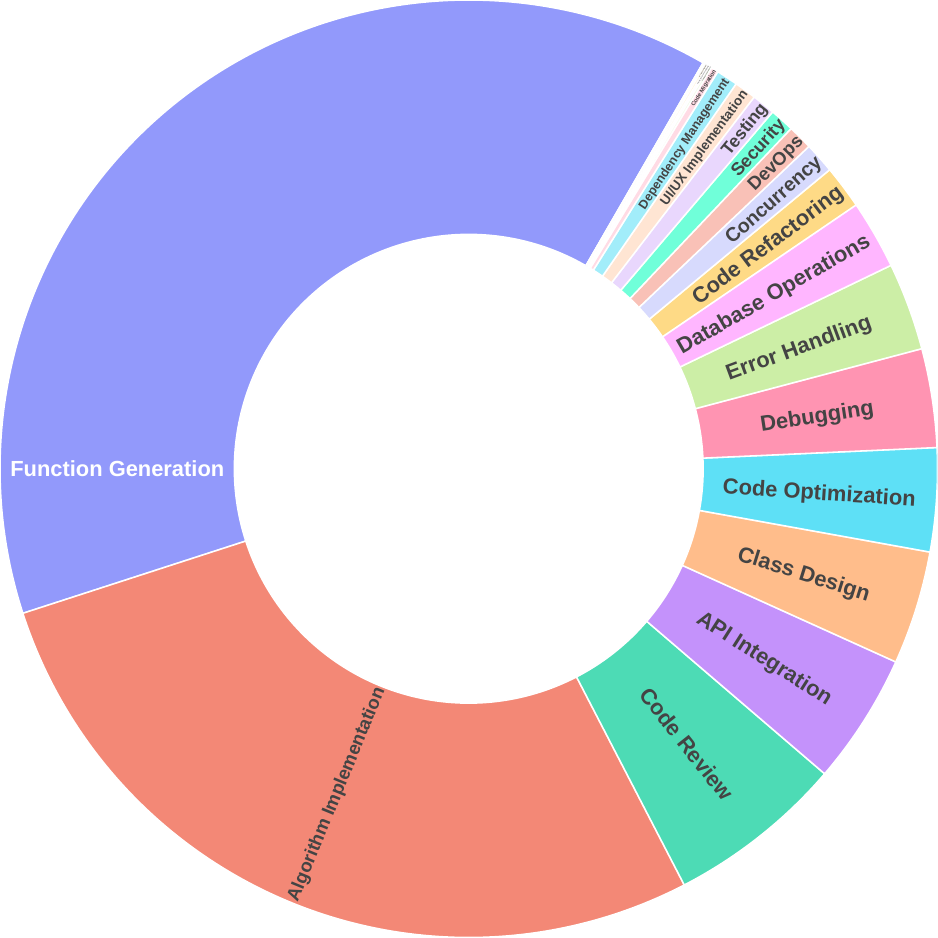}
    \caption{Task categories of the collected Magpie Coding data before filtering.}
    \label{fig:instruction-tags}
\end{figure}

To create the Filter subset, we first generate data using seven state-of-the-art models: Llama-3.1/3.3-70B-Instruct \cite{dubey2024llama}, Qwen2/2.5-72B-Instruct \cite{qwen2,qwen2.5}, Qwen2.5-Coder-32B-Instruct \cite{hui2024qwen25coder}, Qwen2.5-Math-72B-Instruct \cite{yang2024qwen25mathtechnicalreportmathematical}, and Gemma-2-27b-it \cite{team2024gemma}. We then filter for Python-specific content in both instructions and responses, yielding 186K instances. Using \textit{Llama-3.1-8B-Instruct} as our annotator, we label each instance for quality and difficulty (prompts provided in Appendix \ref{appendix: Prompt Template for Dataset Labeling}). We retain only high-quality instances categorized as "Algorithm Implementation" or "Function Generation". Figure \ref{fig:instruction-tags} shows the distribution of task categories before filtering. The final filtered dataset contains 89K coding questions. We follow Magpie's CC-BY-NC 4.0 license.

\section{More Analysis on \dataname{}}
\label{appendix: more stats analysis}

\subsection{Analysis on MBPP Validation Failures Cases with Self Verification} \label{sec: failed mbpp test}

\begin{figure*}[htpb]
    \centering
    \includegraphics[width=0.95\linewidth]{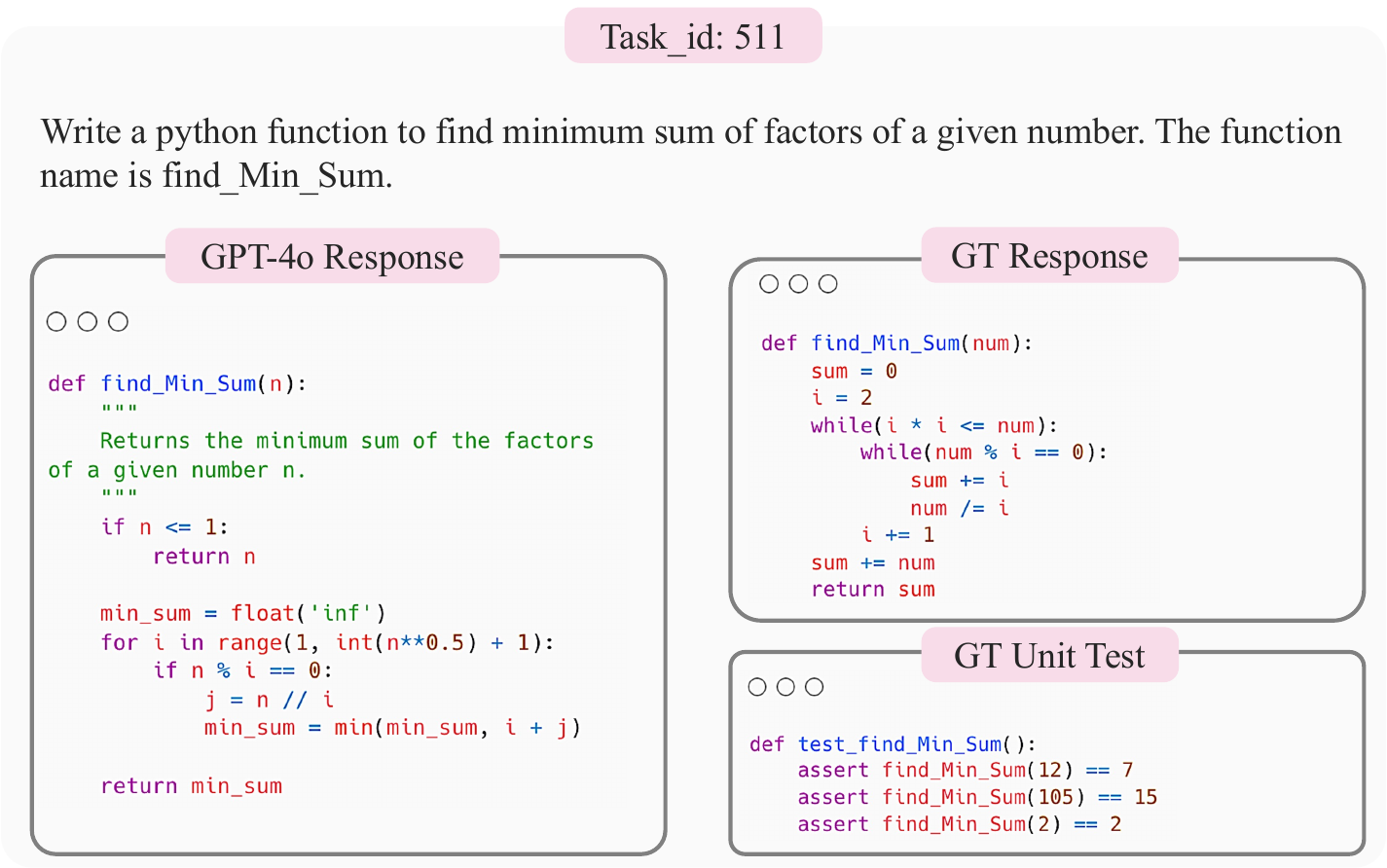}
    \caption{Failed Task 511.}
    \label{fig:mbpp511}
\end{figure*}

\begin{figure*}[htpb]
    \centering
    \includegraphics[width=0.95 \linewidth]{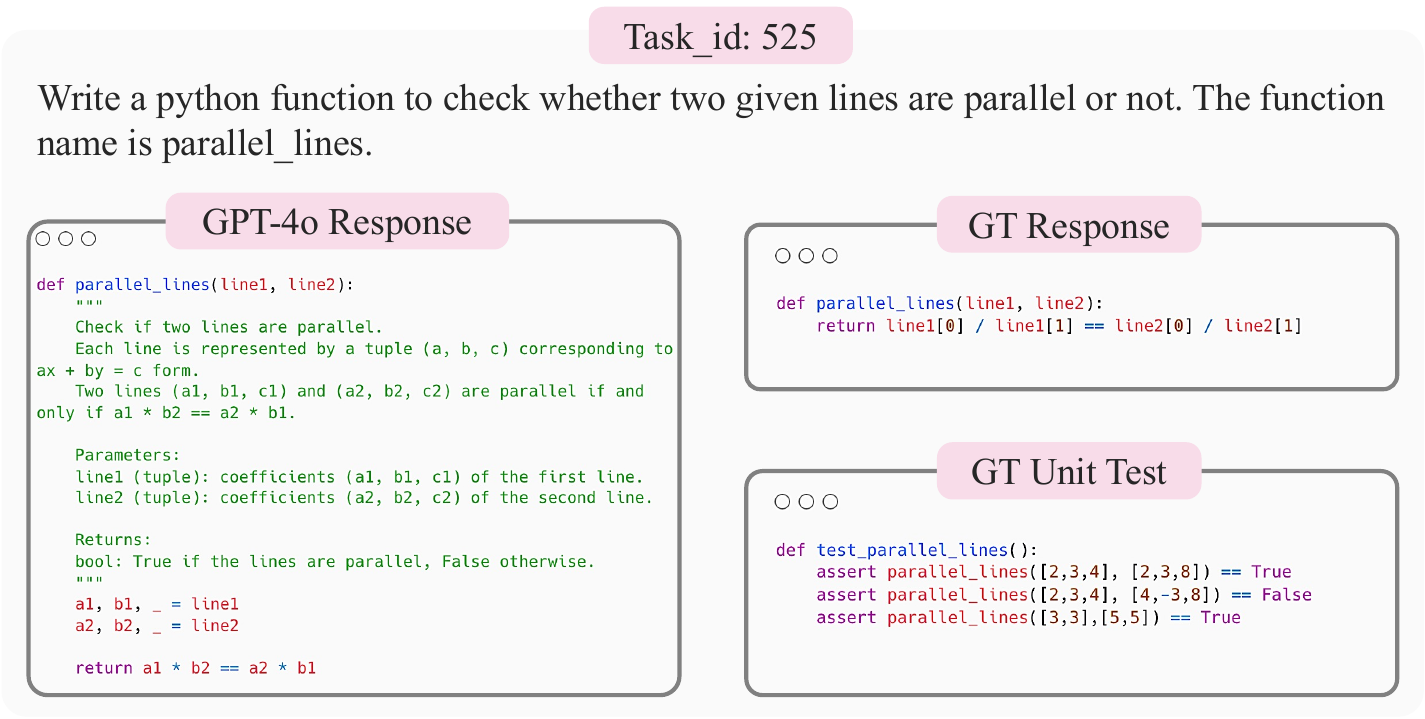}
    \caption{Failed Task 525.}
    \label{fig:mbpp525}
\end{figure*}

Figures \ref{fig:mbpp511} and \ref{fig:mbpp525} illustrate two questions (Task 511 and Task 525) that failed in the unit test during our evaluation on the MBPP validation dataset. In Task 511, which requires computing the minimum sum of factors, the GPT-4o function attempts to minimize the sum of factor pairs by iterating up to the square root of the input number. However, it fails to correctly accumulate all prime factors when multiple factors exist beyond a single pair (e.g., 105, which has factors 3, 5, and 7), leading to incorrect results compared to the ground-truth (GT) implementation, which iteratively divides the number while summing all valid divisors.

In Task 525, the goal is to determine whether two given lines are parallel, with each line represented as a tuple of coefficients. The GT solution verifies parallelism by directly comparing slopes, $\frac{a_1}{b_1} = \frac{a_2}{b_2}$, ensuring compatibility with both the general case of two-element $(a, b)$ tuples and three-element $(a, b, c)$ representations. In contrast, GPT-4o applied an equivalent determinant-based condition, $a_1 \cdot b_2 = a_2 \cdot b_1$, but implicitly assumed that all inputs followed the three-element format, failing to account for the more general case where only $(a, b)$ is provided.  
Due to this assumption, GPT-4o’s implementation produced a mismatch in the third GT unit test case, where the input consists of two-element tuples $(3,3)$ and $(5,5)$, which implicitly represent valid lines with $c=0$.

\subsection{Examples of Dataset Contamination.}
\label{appendix: contamination examples}

We present two examples of instances from \dataname that are similar to those in the MBPP and HumanEval benchmarks. These are illustrated in Figure \ref{Fig: Example 1 of Contaminated Data} and Figure \ref{Fig: Example 2 of Contaminated Data}, respectively.

\begin{figure*}
\begin{tcolorbox}[title=Example 1 of Contaminated Data, promptstyle]
\begin{lstlisting}[breaklines=true, basicstyle=\ttfamily\small]
KodCode: How do I write a Python function to count the number of uppercase letters in a given string?

MBPP/450: Write a python function to count the upper case characters in a given string.

Cosine Similarity: 0.959
\end{lstlisting}
\end{tcolorbox}
\caption{Example 1 of Contaminated Data}
\label{Fig: Example 1 of Contaminated Data}
\end{figure*}

\begin{figure*}
\begin{tcolorbox}[title=Example 2 of Contaminated Data, promptstyle]
\begin{lstlisting}[breaklines=true, basicstyle=\ttfamily\small]
KodCode: def is_prime(n):
    """Returns True if n is a prime number, otherwise False.
    
    >>> is_prime(2) == True
    >>> is_prime(3) == True
    >>> is_prime(5) == True
    >>> is_prime(11) == True
    >>> is_prime(13) == True
    >>> is_prime(0) == False
    >>> is_prime(1) == False
    >>> is_prime(4) == False
    >>> is_prime(6) == False
    >>> is_prime(9) == False
    >>> is_prime(7919) == True
    >>> is_prime(8000) == False
    >>> is_prime(-1) == False
    >>> is_prime(-5) == False
    """

HumanEval/31: def is_prime(n):
    """Return true if a given number is prime, and false otherwise.
    >>> is_prime(6)
    False
    >>> is_prime(101)
    True
    >>> is_prime(11)
    True
    >>> is_prime(13441)
    True
    >>> is_prime(61)
    True
    >>> is_prime(4)
    False
    >>> is_prime(1)
    False

Cosine Similarity: 0.953
\end{lstlisting}
\end{tcolorbox}
\caption{Example 2 of Contaminated Data}
\label{Fig: Example 2 of Contaminated Data}
\end{figure*}

\begin{figure*}[htpb]
\begin{tcolorbox}[title=System Prompt for Fine-Tuning, promptstyle]
Your role as an assistant involves thoroughly exploring questions through a systematic long thinking process before providing the final precise and accurate solutions. This requires engaging in a comprehensive cycle of analysis, summarizing, exploration, reassessment, reflection, backtracing, and iteration to develop well-considered thinking process. Please structure your response into two main sections: Thought and Solution. In the Thought section, detail your reasoning process using the specified format: <think> {thought with steps} </think>. Each step should include detailed considerations such as analisying questions, summarizing relevant findings, brainstorming new ideas, verifying the accuracy of the current steps, refining any errors, and revisiting previous steps. In the Solution section after </think>, synthesize your analysis from the Thought section to present a clear, well-structured solution. Your final answer should be logically organized, precise, and include all essential implementation steps while maintaining conciseness.
\end{tcolorbox}
\caption{System Prompt for Fine-Tuning}
\label{fig: System Prompt for Fine-Tuning}
\end{figure*}
\section{More on Experimental Setups}



\subsection{Supervised Fine-Tuning Setups}
\label{appendix: Instruction-Tuning Setups}

Table \ref{tab: fine-tune hyperparameters} demonstrates the detailed supervised fine-tuning (SFT) hyper-parameters. We perform experiments on a cluster with 768 NVIDIA A100 GPUs. We fine-tune the model using 32 GPUs. The experiments in this paper were conducted using Llama Factory\footnote{\url{https://github.com/hiyouga/LLaMA-Factory}}. For datasets larger than 20K samples, we train for 2 epochs; for datasets smaller than 20K, we increase this to 3 epochs.

\begin{table}[htbp]
\small
\centering
\caption{This table shows the hyper-parameters for supervised fine-tuning.}
\vspace{1em}
\begin{tabular}{ll}
\toprule
\textbf{Hyper-parameter} & \textbf{Value} \\ \midrule
Learning Rate & $1 \times 10^{-5}$ \\
Number of Epochs & $2 (>20K)$ / $3 (<20K)$ \\
Number of Devices & $32$ \\
Per-device Batch Size & $1$ \\
Gradient Accumulation Steps & $4$ \\
Effective Batch Size & $128$ \\
Optimizer & \texttt{Adamw} \\
Learning Rate Scheduler & \texttt{cosine} \\
Warmup Steps & $100$ \\
Max Sequence Length  & $16384$ \\ \bottomrule
\end{tabular}
\label{tab: fine-tune hyperparameters}
\end{table}

We adapt the system prompt from \cite{sky_t1_2025}, modifying it to better accommodate R1-style long chain-of-thought responses during fine-tuning. The full prompt can be found in Figure \ref{fig: System Prompt for Fine-Tuning}.
\section{Prompt Templates}

\subsection{Prompt Template for Dataset Labeling}
\label{appendix: Prompt Template for Dataset Labeling}

Figure \ref{fig: Prompt Template for Labeling Question Quality} and Figure \ref{fig: Prompt Template for Labeling Task Category} detail the prompts for labeling question quality and task category.

\begin{figure*}
\begin{tcolorbox}[title=Prompt Template for Labeling Question Quality, promptstyle]
\lstset{
    basicstyle=\normalfont\sffamily\footnotesize,
    breaklines=true,
    frame=none,
    columns=fullflexible,
}
\begin{lstlisting}[breaklines=true, basicstyle=\ttfamily\small]
# Instruction

You need to rate the quality of the code-related query based on its clarity, specificity, and completeness.

The rating scale is as follows:

- very poor: The query lacks critical code context (e.g., no code samples, error messages, or specific requirements). The problem description is vague or incoherent.
- poor: The query provides incomplete code context or unclear requirements. Important details like programming language, error messages, or expected behavior are missing.
- average: The query includes basic code context and requirements but may need clarification on specific behaviors, edge cases, or implementation details.
- good: The query provides clear code samples, specific requirements, and sufficient context (e.g., language version, environment, expected behavior). Minor details might be missing.
- excellent: The query is comprehensive with complete code examples, clear requirements, relevant error messages if applicable, expected behavior, and implementation constraints. All necessary context is provided for solving the problem.

If the query is not related to code, please rate it as `very poor`. If the query is short but clearly demonstrates the user's intent, please rate it as `good`.

If the query only contains code with no specific instructions, please rate it as `very poor`.

If the query explicitly asks to use programming language other than Python, please rate it as `very poor`.

## User Query
```
{input}
```

## Output Format
Given the user query, you first need to give an assesement, highlighting the strengths and/or weaknesses of the user query.
Then, you need to output a rating from very poor to excellent by filling in the placeholders in [...]:
```
{{   
    "explanation": "[...]",
    "input_quality": "[very poor/poor/average/good/excellent]"
}}
```
\end{lstlisting}
\end{tcolorbox}
\caption{Prompt Template for Labeling Question Quality.}
\label{fig: Prompt Template for Labeling Question Quality}
\end{figure*}

\begin{figure*}
\begin{tcolorbox}[title=Prompt Template for Labeling Task Category, promptstyle]
\lstset{
    basicstyle=\normalfont\sffamily\footnotesize,
    breaklines=true,
    frame=none,
    columns=fullflexible,
}
\begin{lstlisting}[breaklines=true, basicstyle=\ttfamily\small]
# Instruction

Please label the task tags for the user query.

## User Query
```
{input}
```

## Tagging the user input
Please label the task tags for the user query. You will need to analyze the user query and select the most relevant task tag from the list below.

all_task_tags = [
    "Debugging",  # Users ask for help with debugging code. The user should provide the code and error message.
    "Code Review",  # Users ask for help with reviewing code. The user should provide the code.
    "Code Refactoring",  # Users ask for help with refactoring code to improve its structure.
    "Code Optimization",  # Users ask for help with improving code performance or efficiency.
    "Function Generation",  # Users ask for help with creating new functions based on requirements.
    "Class Design",  # Users ask for help with designing classes and object-oriented structures.
    "Algorithm Implementation",  # Users ask for help with implementing specific algorithms or data structures.
    "API Integration",  # Users ask for help with integrating third-party APIs or services.
    "Database Operations",  # Users ask for help with database queries, schema design, or operations.
    "Testing",  # Users ask for help with unit tests, integration tests, or test strategies.
    "Security",  # Users ask for help with security-related implementations or best practices.
    "Error Handling",  # Users ask for help with implementing error handling and validation.
    "Concurrency",  # Users ask for help with multi-threading, async programming, or parallel processing.
    "UI/UX Implementation",  # Users ask for help with frontend implementations or user interface code.
    "DevOps",  # Users ask for help with deployment, CI/CD, or infrastructure code.
    "Documentation",  # Users ask for help with code documentation or technical writing.
    "Dependency Management",  # Users ask for help with package management or dependency issues.
    "Code Migration",  # Users ask for help with migrating code between languages or frameworks.
    "Performance Profiling",  # Users ask for help with identifying and resolving performance bottlenecks.
    "Others"  # Any queries that do not fit into the above categories.
]

## Output Format:
Note that you can only select a single primary tag. Other applicable tags can be added to the list of other tags.
Now, please output your tags below in a json format by filling in the placeholders in <...>:
```
{{ 
    "primary_tag": "<primary tag>",
    "other_tags": ["<tag 1>", "<tag 2>", ... ]
}}
```
\end{lstlisting}
\end{tcolorbox}
\caption{Prompt Template for Labeling Task Category.}
\label{fig: Prompt Template for Labeling Task Category}
\end{figure*}

\subsection{Prompt Template for \textsc{Magpie}-Prefill}
\label{appendix: magpie-prefill template}

We utilize the following prompt template to query the \textit{Qwen2.5-Coder-7B-Instruct} model. We implement the following three different prefillings and replace \{Prefilling Content\} in the template with one of the following options: (1) Write a function to, (2) Write a Python function, (3) Create a function that.

\begin{tcolorbox}[title=Prompt Template for \textsc{Magpie}-Prefill, promptstyle]
\begin{lstlisting}[breaklines=true, basicstyle=\ttfamily\small]
<|im_start|>system
You are Qwen, created by Alibaba Cloud. You are a helpful assistant. You are designed to provide helpful, step-by-step guidance on coding problems. The user will ask you a wide range of coding questions.
Your purpose is to assist users in understanding coding concepts, working through code, and arriving at the correct solutions.<|im_end|>
<|im_start|>user
{Prefilling Content}
\end{lstlisting}
\end{tcolorbox}

\begin{figure}[htpb]
    \centering
    \includegraphics[width=0.9\linewidth]{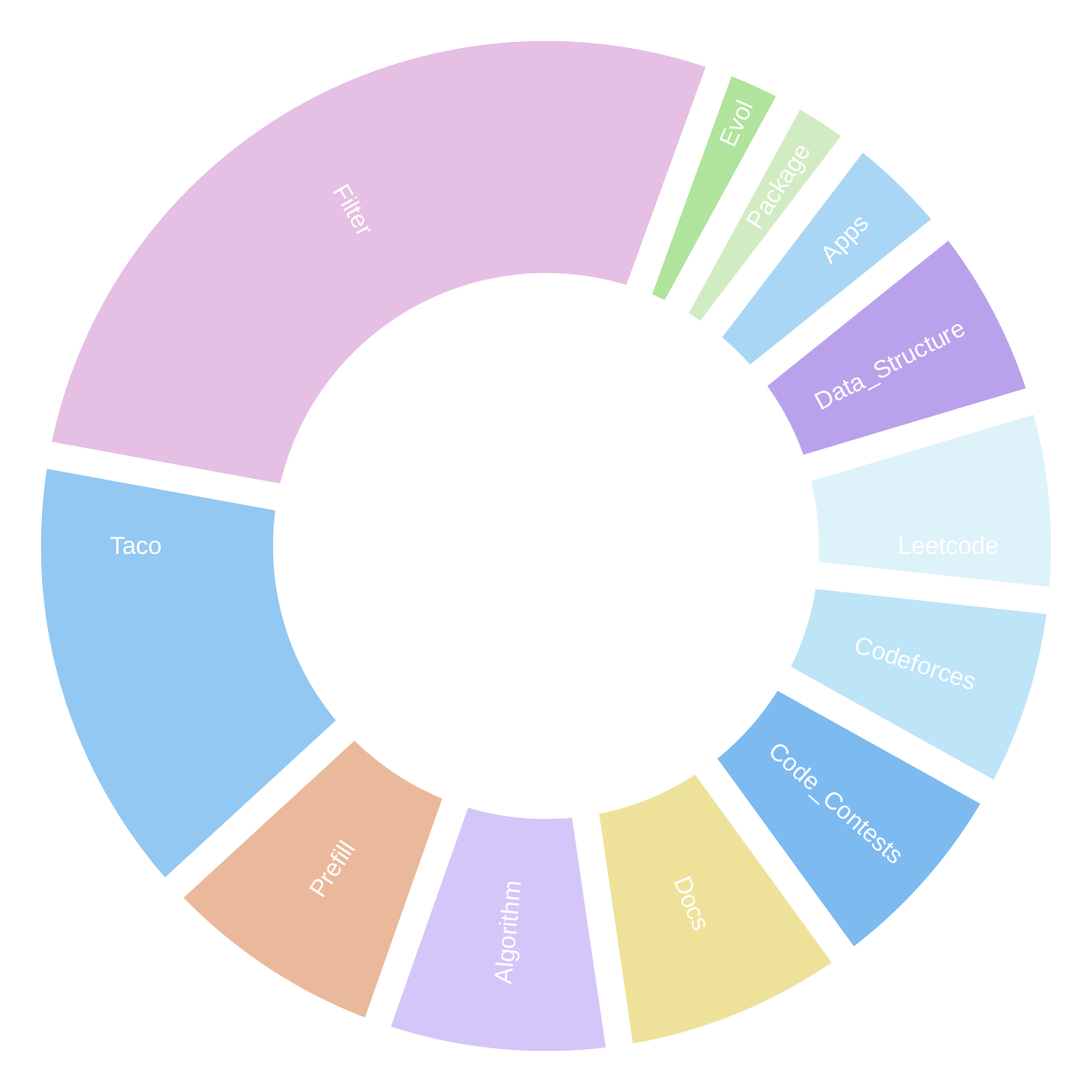}
    \caption{Subset Distribution of \datanamen.}
    \label{fig: subset distribution}
\end{figure}

\subsection{Prompt Template for Generating Coding Assessment Questions}
\label{appendix: code assessment template}

Figure \ref{fig: Prompt Template for Generating Coding Assessment Questions} demonstrates the prompt template for generating coding assessment questions.

\begin{figure*}
\begin{tcolorbox}[title=Prompt Template for Generating Coding Assessment Questions, promptstyle]
\lstset{
    basicstyle=\normalfont\sffamily\footnotesize,
    breaklines=true,
    frame=none,
    columns=fullflexible,
}
\begin{lstlisting}[breaklines=true, basicstyle=\ttfamily\small]
## Task
Design a **Coding Assessment Question**.

## Objective
Analyze the provided sample questions and create an additional question that aligns with the existing set in terms of style, complexity, and scope.

## Guidelines

### Question Style
- Carefully examine the format and presentation of the given questions.
- Maintain a consistent tone and language used in the original set.

### Question Length
- Ensure your new question is of comparable length to the existing ones.
- If the original questions vary in length, aim for the average length among them.

### Difficulty Level
- Assess the cognitive and technical challenges presented in the sample questions.
- Match the complexity of concepts, algorithms, or programming techniques required.

### Topic Alignment
- Identify the core programming concepts or domains covered in the existing questions.
- Create a question that explores a related or complementary area within the same general topic.

### Question Uniqueness
- While maintaining similarity, ensure your new question is not a mere rephrasing of an existing one.
- Introduce a novel problem or scenario that tests the same skills in a different context.

## Output
Create a new question that matches the style of the existing set of questions. Output one new question only. Direct output the question, without adding question number or any other text in the beginning.

## Question 1 
{Seed 1}

## Question 2 
{Seed 2}

## Question 3 
{Seed 3}

## Question 4 

\end{lstlisting}
\end{tcolorbox}
\caption{Prompt Template for Generating Coding Assessment Questions}
\label{fig: Prompt Template for Generating Coding Assessment Questions}
\end{figure*}

\subsection{Prompt Template for Converting DSA Code Snippets to Questions}
\label{appendix: algorithm template}

Figure \ref{fig: Prompt Template for Algorithm Subset} demonstrates the prompt template for generating questions for the algorithm subset from the code \& algorithm snippets.

\begin{figure*}
\vspace{-2em}
\begin{tcolorbox}[title=Prompt Template for Algorithm Subset, promptstyle]
\lstset{
    basicstyle=\normalfont\sffamily\footnotesize,
    breaklines=true,
    frame=none,
    columns=fullflexible,
}
\begin{lstlisting}[breaklines=true, basicstyle=\ttfamily\scriptsize]

## Task
Design a **Coding Assessment Question**.

## Objective
As an experienced programming instructor, you are designing programming questions to assess students' understanding of algorithms and data structures provided in the code snippets. Your question should require students to write code that demonstrates their comprehension of fundamental and advanced concepts from the code snippets.

## Guidelines

### Algorithm Analysis
Provide a thorough analysis of the algorithm or data structure provided in the code snippets, including but not limited to the following aspects:

#### Core Identification
* **Algorithm/Data Structure**: State the name, type (e.g., sorting algorithm, tree data structure), and main purpose.
* **Complexity**: Outline time and space complexity.
* **Principles**: Summarize its core operational steps or key mechanisms.

#### Characteristics & Applications
* **Properties**: Highlight essential properties (e.g., sorting stability, traversal order).
* **Common Use Cases**: Describe typical scenarios where this algorithm or structure is effective.
* **Strengths/Limitations**: Identify its key advantages and drawbacks, and specify when it's most suitable to use or avoid.

#### Implementation Challenges
* **Edge Cases**: Describe common edge cases students should consider.
* **Performance Bottlenecks**: Identify any parts that can slow down execution or use excess memory.
* **Error Scenarios**: Explain situations that might lead to incorrect results if not handled correctly.
* **Optimization Points**: Note potential improvements or alternatives to enhance performance.

### Question Style
Based on the above analysis, craft a question that is:
* **Challenging** and requires a well-thought-out coding solution.
* **Clear and self-contained**, with all necessary information for solving it provided.
* **Focused on function implementation**, specifying:
  * Expected **input and output formats**.
  * Any **constraints or limitations**.
  * **Performance requirements**, if applicable.
* Enriched with a brief **scenario or context** (if it enhances clarity).
* Thoroughly examining the algorithm or data structure without referencing any example code directly.

## Output Format
Please use the following output format for consistency.

<|Analysis Begin|>

[Write your analysis here]

<|Analysis End|>

<|Question Begin|>

[Write the coding question here]

<|Question End|>

## Code Snippets


\end{lstlisting}
\end{tcolorbox}
\caption{Prompt Template for Algorithm Subset}
\label{fig: Prompt Template for Algorithm Subset}
\end{figure*}

\subsection{Prompt Template for Converting Technical Documentations to Coding Questions}
\label{appendix: docs template}

Figure \ref{fig: Prompt Template for Docs Subset} demonstrates the prompt template for generating questions for the Docs subset.

\begin{figure*}
\vspace{-2em}
\begin{tcolorbox}[title=Prompt Template for Docs Subset, promptstyle]
\lstset{
    basicstyle=\normalfont\sffamily\footnotesize,
    breaklines=true,
    frame=none,
    columns=fullflexible,
}
\begin{lstlisting}[breaklines=true, basicstyle=\ttfamily\small]
## Task
Design a **Coding Assessment Question**.

## Objective
As an experienced programming instructor, you are designing programming questions to assess students' understanding of {PACKAGE_NAME}. Your question should require students to write code that demonstrates their comprehension of fundamental and advanced concepts of this package.

## Guidelines

I will provide you with the documentation of {PACKAGE_NAME}. It could be a jupyter notebook, txt, or rst file. Please use it to design the question. 

You should first analyze the documentation provided and understand the design of the package. Then, craft a question that is:
* **Challenging** and requires a well-thought-out coding solution.
* **Clear and self-contained**, with all necessary information for solving it provided.
* **Focused on function implementation**, specifying:
  * Expected **input and output formats**.
  * Any **constraints or limitations**.
  * **Performance requirements**, if applicable.

## Output Format
Please use the following output format for consistency.

<|Analysis Begin|>

[Write your analysis here]

<|Analysis End|>

<|Question Begin|>

[Write the coding question here. If you believe this document is not enough to design a question, please output "BAD_DOCUMENT" in this section.]

<|Question End|>

## Documentation
{CONTENT}

----------------
Now, please output the analysis and the question.

\end{lstlisting}
\end{tcolorbox}
\caption{Prompt Template for Docs Subset}
\label{fig: Prompt Template for Docs Subset}
\end{figure*}

\subsection{Prompt Template for Generating Solutions and Tests}
\label{appendix: solution template}

Figure \ref{fig: Prompt Template for Generating Solutions and Tests} demonstrates the prompt template for generating solutions and corresponding unit tests using GPT-4o-0513.

\begin{figure*}
\begin{tcolorbox}[title=Prompt Template for Generating Solutions and Tests, promptstyle]
\lstset{
    basicstyle=\normalfont\sffamily\footnotesize,
    breaklines=true,
    frame=none,
    columns=fullflexible,
}
\begin{lstlisting}[breaklines=true, basicstyle=\ttfamily\small]
## Task:
Please Answer the question and generate unit tests to verify your answer.

## Output Format:
Your solution and unit tests should be presented in markdown Python code format within the specified sections below. Ensure your code is within code blocks. For the tests, use pytest style by defining individual test functions (without classes) and using assert statements. 

<|Solution Begin|>
[Solution Code in Python]
<|Solution End|>
<|Test Begin|>
[Unit Test Code in Python]
<|Test End|>

## Example
Below is an example output format implementing a simple a + b function.

<|Solution Begin|>
```python
def add(a, b):
    """
    Returns the sum of a and b.
    """
    return a + b
```
<|Solution End|>

<|Test Begin|>
```python
from solution import add

def test_add_positive_numbers():
    assert add(2, 3) == 5

def test_add_with_zero():
    assert add(0, 5) == 5
    assert add(5, 0) == 5

def test_add_negative_numbers():
    assert add(-1, -1) == -2

def test_add_mixed_sign_numbers():
    assert add(-1, 3) == 2
```
<|Test End|>

## Question:

\end{lstlisting}
\end{tcolorbox}
\caption{Prompt Template for Generating Solutions and Tests}
\label{fig: Prompt Template for Generating Solutions and Tests}
\end{figure*}

\subsection{Prompt Template for Style Converter}
\label{appendix: style converter template}

Figure \ref{fig: Prompt Template for Style Converter} demonstrates the prompt template for converting question styles.

\begin{figure*}
\vspace{1em}
\begin{tcolorbox}[title=Prompt Template for Style Converter, promptstyle]
\lstset{
    basicstyle=\normalfont\sffamily\footnotesize,
    breaklines=true,
    frame=none,
    columns=fullflexible,
}
\begin{lstlisting}[breaklines=true, basicstyle=\ttfamily\scriptsize]
## Task:
Please convert the given coding task to a coding completion task.

## Instruction

I will give you a coding task, and you goal is to convert it to a completion task. For example, if the coding task is "Find the longest common prefix among a list of strings.", the completion should be a task requiring the user to write a function that takes a list of strings as input and returns the longest common prefix as shown in the example below. Please ensure that the completion task contains the function definition, necessary imports, description, and test cases (if applicable). 

I will provide you a coding task, along with a unit test and a solution. You can refer to them for the test cases and the function definition, but do not put the solution in the completion task.

Coding Task: Find the longest common prefix among a list of strings.

Unit Test:
```python
from solution import longest_common_prefix

def test_longest_common_prefix():
    assert longest_common_prefix(["flower", "flow", "flight"]) == "fl"
    assert longest_common_prefix(["dog", "racecar", "car"]) == ""
```

Solution:
```python
def longestCommonPrefix(strs: List[str]) -> str:
    # Sort the list of strings
    strs.sort()
    
    # Only need to compare first and last strings after sorting
    first = strs[0]
    last = strs[-1]
    
    # Find the common prefix between first and last
    i = 0
    while i < min(len(first), len(last)) and first[i] == last[i]:
        i += 1
    
    return first[:i]
```

Completion Task you should generate:
```python
def longest_common_prefix(strs: List[str]) -> str:
    """ Find the longest common prefix among a list of strings.
    >>> longest_common_prefix(["flower", "flow", "flight"]) "fl"
    >>> longest_common_prefix(["dog", "racecar", "car"]) ""
    """
```

Now, I will give you a coding task, and you goal is to convert it to a completion task.
## Input Information

Coding Task: [Coding Task Placeholder]

Unit Test: [Unit Test Placeholder]

Solution: [Solution Placeholder]

## Output Format:
<|Completion Begin|>
[Completion Task in Python]
<|Completion End|>

Please output the completion task strictly in the provided format, without adding any other information.

\end{lstlisting}
\end{tcolorbox}
\caption{Prompt Template for Style Converter}
\vspace{1em}
\label{fig: Prompt Template for Style Converter}
\end{figure*}
\section{\dataname{} Examples}

We present representative examples from \dataname{}~subsets in Figure \ref{Example of datanamen Subset: Prefill} to Figure \ref{Example of datanamen Subset: Filter}. The distribution of each subset is demonstrated in Figure \ref{fig: subset distribution}.

\begin{figure*}
\begin{tcolorbox}[title=\datanamen~Subset: Prefill, promptstyle]
\begin{lstlisting}[breaklines=true, basicstyle=\ttfamily\scriptsize]
Write a function to find max and min from an array in Python. I am looking for an O(n) time complexity solution.
\end{lstlisting}
\end{tcolorbox}
\caption{Example of \datanamen~Subset: Prefill}
\vspace{-1em}
\label{Example of datanamen Subset: Prefill}
\end{figure*}

\begin{figure*}
\begin{tcolorbox}[title=\datanamen~Subset: Package, promptstyle]
\begin{lstlisting}[breaklines=true, basicstyle=\ttfamily\scriptsize]
Design and implement a function that, given a list of integers, returns a new list where each element is the product of all other elements in the list except the one at the current index. You should solve this without using division and in O(n) time complexity.

The function should output:
- list: A list where each element is the product of all other elements in the input list except the element at the current index.

You should write self-contained code starting with:
```
def product_except_self(nums):
```
The first example from KodCode: Package: You are tasked with creating a function that calculates the similarity between two text documents based on the Jaccard similarity coefficient. The Jaccard similarity coefficient is defined as the size of the intersection divided by the size of the union of the sample sets. This function should take two strings as input, tokenize them into words, and then compute the similarity.

#### Function Specification:

**Function Name**: `jaccard_similarity`

**Parameters**:
- `doc1` (str): The first document as a string.
- `doc2` (str): The second document as a string.

**Behavior**:
1. Tokenize both input strings into sets of words.
2. Compute the intersection of these sets.
3. Compute the union of these sets.
4. Calculate the Jaccard similarity coefficient: (size of intersection) / (size of union).
5. Return the Jaccard similarity coefficient as a float.

**Example**:
```python
def jaccard_similarity(doc1, doc2):
    # Tokenize the documents
    set1 = set(doc1.split())
    set2 = set(doc2.split())
    
    # Compute intersection and union
    intersection = set1.intersection(set2)
    union = set1.union(set2)
    
    # Calculate and return Jaccard similarity coefficient
    return len(intersection) / len(union)

# Example usage
doc1 = "the cat in the hat"
doc2 = "the cat with a hat"
print(jaccard_similarity(doc1, doc2))  # Output: 0.5
```
\end{lstlisting}
\end{tcolorbox}
\caption{Example of \datanamen~Subset: Package}
\label{Example of datanamen Subset: Package}
\end{figure*}

\begin{figure*}
\begin{tcolorbox}[title=\datanamen~Subset: Codeforces, promptstyle]
\begin{lstlisting}[breaklines=true, basicstyle=\ttfamily\scriptsize]
Many citizens of Gridland have taken up gardening as a hobby. Each gardener has a rectangular plot represented by a grid with $$$N$$$ rows and $$$M$$$ columns. Some cells in the grid may already contain flowers. Each gardener wants to plant saplings in the remaining empty cells such that no two saplings are adjacent to each other, neither horizontally, vertically, nor diagonally. The goal is to determine the maximum number of saplings that can be planted in the given grid.

### Input
- The first line of input contains two integers, $$$N$$$ and $$$M$$$ ($$$1 \leq N, M \leq 1000$$$), representing the dimensions of the grid.
- The next $$$N$$$ lines each contain $$$M$$$ characters, representing the grid. Each character is either a 'F' (which means a flower is already planted in that cell) or an 'E' (which means the cell is empty).

### Output
- Output a single integer, the maximum number of saplings that can be planted.

### Example

#### Input
```
3 4
E E E E
E F E E
E E E E
```

#### Output
```
4
```

### Explanation
The optimal arrangement of planting saplings would be:

```
S E S E
E F E E
S E S E
```

Placing saplings ('S') in this way ensures no two saplings are adjacent and maximizes the number of saplings planted, which totals to 4 in this example.
\end{lstlisting}
\end{tcolorbox}
\caption{Example of \datanamen~Subset: Codeforces}
\label{Example of datanamen Subset: Codeforces}
\vspace{-2em}
\end{figure*}

\begin{figure*}
\vspace{-1em}
\begin{tcolorbox}[title=\datanamen~Subset: Leetcode, promptstyle]
\begin{lstlisting}[breaklines=true, basicstyle=\ttfamily\scriptsize]
Given a list of integers `nums`, find the maximum product of any two distinct elements in the list. Return the maximum product. For example, given `nums = [5, 3, -1, 9, -7]`, the maximum product would be `5 * 9 = 45`.
\end{lstlisting}
\end{tcolorbox}
\caption{Example of \datanamen~Subset: Leetcode}
\label{Example of datanamen Subset: Leetcode}
\end{figure*}

\begin{figure*}
\begin{tcolorbox}[title=\datanamen~Subset: Apps, promptstyle]
\begin{lstlisting}[breaklines=true, basicstyle=\ttfamily\scriptsize]
## Task
Design a **Matrix Multiplication Function**.

## Objective
Write a function that multiplies two matrices of compatible dimensions and returns the result. Your implementation should include a check for dimension compatibility before attempting to multiply the matrices.

## Guidelines

### Function Signature
```python
def matrix_multiply(A, B):
    pass
```

### Input
- `A`: A list of lists representing matrix A, where each inner list represents a row of the matrix.
- `B`: A list of lists representing matrix B, where each inner list represents a row of the matrix.

### Output
- A new matrix representing the product of matrices A and B.

### Constraints
- The number of columns in matrix A should be equal to the number of rows in matrix B to be compatible for multiplication.
- Both matrices can be empty, and the function should return an empty matrix in such cases.

### Example
```python
A = [
    [1, 2, 3],
    [4, 5, 6]
]

B = [
    [7, 8],
    [9, 10],
    [11, 12]
]

print(matrix_multiply(A, B))
# Output: [[58, 64], [139, 154]]
```

### Explanation
The product of matrices A and B results in a new matrix where each element at position (i, j) is calculated as the sum of element-wise products of row i from matrix A and column j from matrix B.

### Constraints
1. Matrices A and B are rectangular.
2. Elements of matrices A and B can be any real numbers.

### Notes
- If the dimensions of the two matrices are incompatible, raise a ValueError with a message "Incompatible dimensions for matrix multiplication".
- The function should handle matrices with any number of rows and columns, but keep in mind the complexity of your solution, as matrix multiplication can be time-consuming for very large matrices.
\end{lstlisting}
\end{tcolorbox}
\caption{Example of \datanamen~Subset: Apps}
\label{Example of datanamen Subset: Apps}
\end{figure*}

\begin{figure*}
\begin{tcolorbox}[title=\datanamen~Subset: Algorithm, promptstyle]
\begin{lstlisting}[breaklines=true, basicstyle=\ttfamily\scriptsize]
You have been hired to develop a real-time weather dashboard that fetches and displays current weather data for a given list of cities. Your task is to write a function that processes and displays weather data from the OpenWeatherMap API for a specified list of cities. You need to ensure robust handling of potential issues such as missing or malformed data. You will implement the following function:

```python
def get_and_display_weather(cities: list) -> None:
    """
    Fetches current weather data from the OpenWeatherMap API for a specified list of cities and displays it in a formatted table.
    
    The API endpoint to use is:
    "http://api.openweathermap.org/data/2.5/weather?q={city}&appid={API_KEY}"

    Requirements:
    - Handle missing or malformed data gracefully, ensuring the program does not crash.
    - Display weather information in degrees Celsius.
    
    The table should display the following columns:
    - City
    - Weather (e.g., clear sky, rain)
    - Temperature
    - Humidity (%)
    - Wind Speed (m/s)
    
    Example of the table format:
    ```
    +-------------+------------+-------------+-----------+------------+
    | City        | Weather    | - Temperature
 | Humidity (%) | Wind Speed (m/s) |
    +-------------+------------+-------------+-----------+------------+
    | New York    | Clear sky  | 22          | 55        | 5          |
    | Los Angeles | Cloudy     | 18          | 60        | 3          |
    | ...                                                           |
    +-------------+------------+-------------+-----------+------------+
    ```

    Parameters:
    - cities: list of strings representing the city names.

    """
    pass
```

### Input/Output Format

- **Input**: A list of city names (e.g., ["New York", "Los Angeles", "Mumbai"]).
- **Output**: The function should output the formatted table with the columns specified above.
  
### Constraints

- Do not assume the API will always return all required fields for each city.
- Consider edge cases where some fields might be missing or improperly formatted.
- Ensure that the temperature conversion to Celsius is accurate.
  
### Performance Requirements

- Efficient handling of network requests and JSON parsing.
- Ensure the table can handle multiple cities without significant delay in processing time.
\end{lstlisting}
\end{tcolorbox}
\caption{Example of \datanamen~Subset: Algorithm}
\label{Example of datanamen Subset: Algorithm}
\end{figure*}

\begin{figure*}
\begin{tcolorbox}[title=\datanamen~Subset: Filter, promptstyle]
\begin{lstlisting}[breaklines=true, basicstyle=\ttfamily\scriptsize]
Given a string, find the longest palindromic subsequence in that string.

Example:
Input: "banana"
Output: "anana"

Example:
Input: "abcd"
Output: "" (because there's no palindromic subsequence)

Note: A subsequence is a sequence that appears in the same relative order, but not necessarily contiguous.
\end{lstlisting}
\end{tcolorbox}
\caption{Example of \datanamen~Subset: Filter}
\label{Example of datanamen Subset: Filter}
\end{figure*}

\end{document}